\pdfoutput=1

\documentclass[11pt]{article}
\usepackage{authblk}

\usepackage[]{acl}

\usepackage{booktabs}
\usepackage{graphicx}
\usepackage{enumitem}
\usepackage{xcolor}
\usepackage{multicol}
\usepackage{amsmath}
\usepackage{bold-extra}
\usepackage{subcaption}
\usepackage{float}
\usepackage{nicefrac}
\usepackage{color, colortbl}
\usepackage{times}
\usepackage{latexsym}
\usepackage{xcolor}
\usepackage{amssymb} 
\usepackage{multirow}
\usepackage{tablefootnote}

\usepackage{textcomp}
\usepackage{amssymb}
\usepackage{pifont}
\renewcommand{\footnotesize}{\fontsize{8.5pt}{10.5pt}\selectfont}

\usepackage[T1]{fontenc}
\usepackage[utf8]{inputenc}

\usepackage{microtype}

\renewcommand\footnotemark{}

\title{\textsc{EntityCS}: Improving Zero-Shot Cross-lingual Transfer\\with Entity-Centric Code Switching}

\author[1,2,*]{Chenxi Whitehouse\thanks{$^*$Work conducted as Research Intern at Huawei Noah’s Ark Lab, London, UK.}}
\author[2]{Fenia Christopoulou}
\author[2]{Ignacio Iacobacci}
\affil[1]{City, University of London}
\affil[2]{Huawei Noah's Ark Lab, London, UK}
\affil[ ]{\tt chenxi.whitehouse@city.ac.uk}
\affil[ ]{\tt \{efstathia.christopoulou, ignacio.iacoboacci\}@huawei.com}

\begin{document}

\maketitle
\begin{abstract}
Accurate alignment between languages is fundamental for improving cross-lingual pre-trained language models (XLMs). Motivated by the natural phenomenon of code-switching (CS) in multilingual speakers, CS has been used as an effective data augmentation method that offers language alignment at the word- or phrase-level, in contrast to sentence-level via parallel instances. 
Existing approaches either use dictionaries or parallel sentences with word alignment to generate CS data by randomly switching words in a sentence.
However, such methods can be suboptimal as dictionaries disregard semantics, and syntax might become invalid after random word switching.
In this work, we propose \textsc{EntityCS}, a method that focuses on \textsc{Entity}-level \textsc{C}ode-\textsc{S}witching to capture fine-grained cross-lingual semantics without corrupting syntax. 
We use Wikidata and English Wikipedia to construct an entity-centric CS corpus by switching entities to their counterparts in other languages.
We further propose entity-oriented masking strategies during intermediate model training on the \textsc{EntityCS} corpus for improving entity prediction.
Evaluation of the trained models on four entity-centric downstream tasks shows consistent improvements over the baseline with a notable increase of 10\% in Fact Retrieval. 
We release the corpus and models to assist research on code-switching and enriching XLMs with external knowledge\footnote{Code and models are available at \url{https://github.com/huawei-noah/noah-research/tree/master/NLP/EntityCS}.}.

\end{abstract}

\begin{figure}[t!]
\centering
    \includegraphics[width=\linewidth]{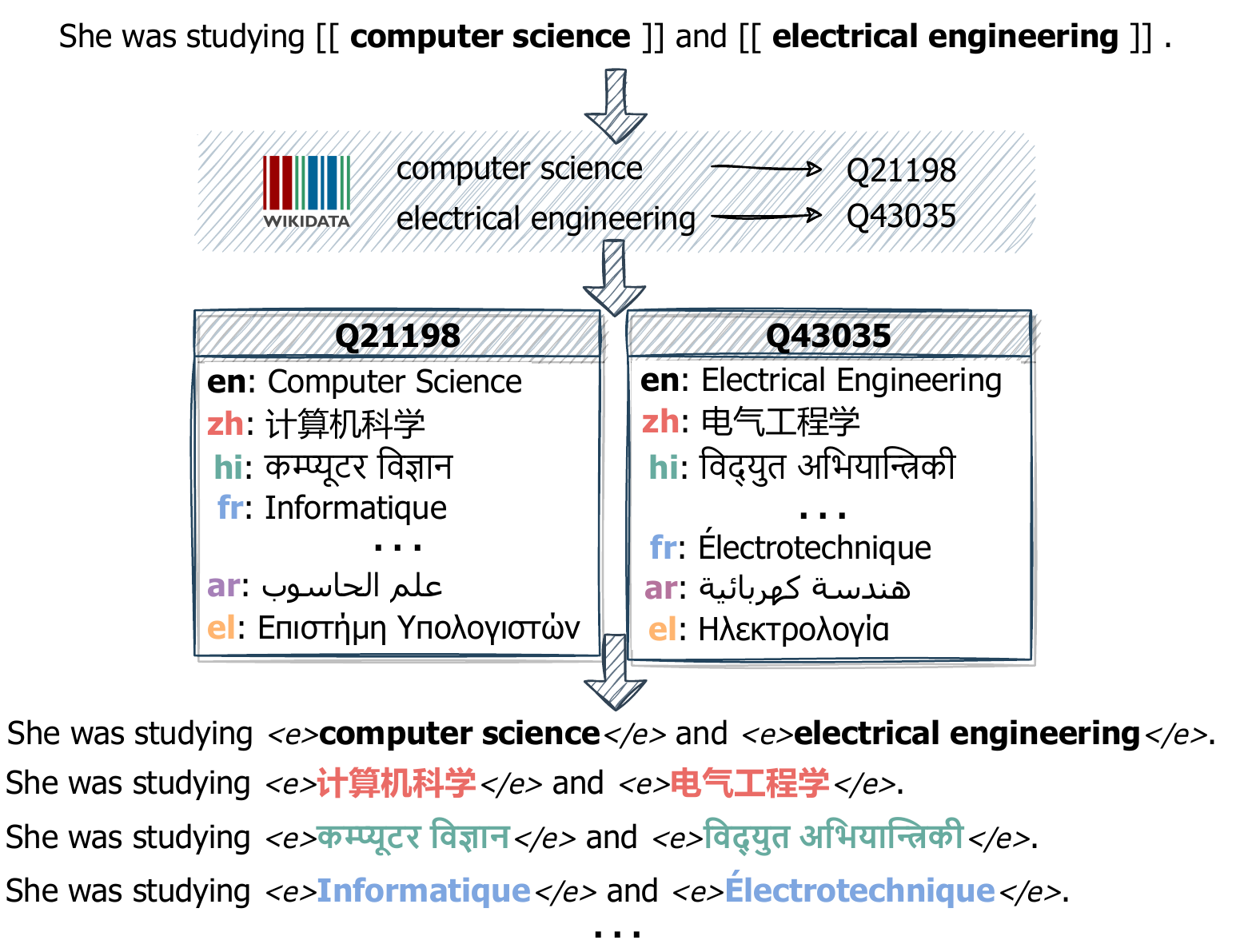}
    \caption{Illustration of generating \textsc{EntityCS} sentences from an English sentence extracted from Wikipedia. Entities in double square brackets indicate wikilinks.}
\label{fig:cs_example}
\end{figure}

\section{Introduction}
Cross-lingual pre-trained Language Models (XLMs) such as mBERT~\cite{devlin-etal-2019-bert} and XLM-R~\cite{conneau-etal-2020-unsupervised}, have achieved state-of-the-art zero-shot cross-lingual transferability across diverse Natural Language Understanding (NLU) tasks.
Such models have been particularly enhanced with the use of bilingual parallel sentences together with alignment methods~\citep{Yang2020AlternatingLM, chi-etal-2021-infoxlm,hu-etal-2021-explicit, gritta-iacobacci-2021-xeroalign,feng-etal-2022-language}.
However, obtaining high-quality parallel data is expensive, especially for low-resource languages.
Therefore, alternative data augmentation approaches have been proposed, one of which is Code Switching. 

Code Switching (CS) is a phenomenon when multilingual speakers alternate words between languages when they speak, which has been studied for many years~\cite{gumperz1977sociolinguistic,khanuja-etal-2020-gluecos, dogruoz-etal-2021-survey}.
Code-switched sentences consist of words or phrases in different languages, therefore they capture the semantics of finer-grained cross-lingual expressions compared to parallel sentences, and have been used for multilingual intermediate training~\citep{Yang2020AlternatingLM} and fine-tuning~\citep{Qin2020CoSDAMLMC, krishnan-etal-2021-multilingual}.
Nevertheless, since manually creating large-scale CS datasets is costly and only a few natural CS texts exist~\citep{barik-etal-2019-normalization, xiang-etal-2020-sina, chakravarthi-etal-2020-corpus, ASCEND-corpus}, research has turned to automatic CS data generation.
Some of those approaches generate CS data via dictionaries, usually ignoring ambiguity~\citep{Qin2020CoSDAMLMC,conneau-etal-2020-emerging}. 
Others require parallel data and an alignment method to match words or phrases between languages \citep{Yang2020AlternatingLM,rizvi-etal-2021-gcm}. In both cases, what is switched is chosen randomly, potentially resulting in syntactically odd sentences or switching to words with little semantic content (e.g. conjunctions).

On the contrary, entities contain external knowledge and do not alter sentence syntax if replaced with other entities, which mitigates the need for any parallel data or word alignment tools.
Motivated by this, we propose \textsc{EntityCS}, a \textsc{C}ode-\textsc{S}witching method that focuses on \textsc{Entities}, as illustrated in \autoref{fig:cs_example}.
Resources such as Wikipedia and Wikidata offer rich cross-lingual entity-level information, which has been proven beneficial in XLMs pre-training~\citep{jiang-etal-2020-x,calixto-etal-2021-wikipedia,jiang2022xlm}.
We use such resources to generate an entity-based CS corpus for the intermediate training of XLMs.
Entities in wikilinks\footnote{\url{https://en.wikipedia.org/wiki/Help:Link\#Wikilinks_(internal_links)}} are switched to their counterparts in other languages retrieved from the Wikidata Knowledge Base (KB), thus alleviating ambiguity.

Using the \textsc{EntityCS} corpus, we propose a series of masking strategies that focus on enhancing Entity Prediction (EP) for better cross-lingual entity representations.
We evaluate the models on entity-centric downstream tasks including Named Entity Recognition (NER), Fact Retrieval, Slot Filling (SF) and Word Sense Disambiguation (WSD).
Extensive experiments demonstrate that our models outperform the baseline on zero-shot cross-lingual transfer, with +2.8\% improvement on NER, surpassing the prior best result that uses large amounts of parallel data, +10.0\% on Fact Retrieval, +2.4\% on Slot Filling, and +1.3\% on WSD.

The main contributions of this work include: 
a) construction of an entity-level CS corpus, \textsc{EntityCS}, based solely on the English Wikipedia and Wikidata, mitigating the need for parallel data, word-alignment methods or dictionaries; 
b) a series of intermediate training objectives, focusing on Entity Prediction; 
c) improvement of zero-shot performance on NER, Fact Retrieval, Slot Filling and WSD; 
d) further analysis of model errors, the behaviour of different masking strategies throughout training as well as impact across languages, demonstrating how our models particularly benefit non-Latin script languages.

\begin{table}[t!]
\centering
\scalebox{0.85}{
\begin{tabular}{lr}
\toprule
\textsc{Statistic} & \textsc{Count} \\ \midrule
Languages                     & 93             \\
English Sentences             & 54,469,214     \\
English Entities              & 104,593,076    \\
Average Sentence Length       & 23.37          \\
Average Entities per Sentence & 2  \\ \midrule
CS Sentences per EN Sentence     & $\leq$ 5 \\
CS Sentences       & 231,124,422    \\
CS Entities        & 420,907,878    \\

\bottomrule
\end{tabular}
}
\caption{Statistics of the \textsc{EntityCS} Corpus.}
 \label{tab:cs_stats}
\end{table}

\section{Methodology}
We introduce the details of the \textsc{EntityCS} corpus construction, as well as different entity-oriented masking strategies used in our experiments.

\subsection{\textsc{EntityCS} Corpus Construction} 
\label{corpus_construction}

Wikipedia is a multilingual online encyclopedia available in more than 300 languages\footnote{\url{https://en.wikipedia.org/wiki/Wikipedia}}.
Structured data of Wikipedia articles are stored in Wikidata, a multilingual document-oriented database.
With more than six million articles, English Wikipedia has the potential to serve as a rich resource for generating CS data.
We use English Wikipedia and leverage entity information from Wikidata to construct an entity-based CS corpus.

To achieve this, we make use of wikilinks in Wikipedia, i.e. links from one page to another.
We use the English Wikipedia dump\footnote{\url{https://dumps.wikimedia.org/enwiki/latest/} (Nov 2021 version).} and extract raw text with WikiExtractor\footnote{\url{https://github.com/attardi/wikiextractor}} while keeping track of wikilinks.
Wikilinks are typically surrounded by square brackets in Wikipedia dump, in the format of [[\textit{entity} | \textit{display text}]], where \textit{entity} is the title of the target Wikipedia page it links to, and \textit{display text} corresponds to what is displayed in the current article.
We then employ SpaCy\footnote{\url{https://spacy.io/}} for sentence segmentation.
Since we are interested in creating entity-level CS instances, we only keep sentences containing at least one wikilink.
Sentences longer than 128 words are also removed.
This results in 54.5M English sentences and 104M entities in the final \textsc{EntityCS} corpus.

As illustrated in \autoref{fig:cs_example}, given an English sentence with wikilinks, we first map the entity in each wikilink to its corresponding Wikidata ID and retrieve its available translations from Wikidata.
For each sentence, we check which languages have translations for all entities in that sentence, and consider those as candidates for code-switching.
We select 92 target languages in total, which are the overlap between the available languages in Wikidata and XLM-R \citep{conneau-etal-2020-unsupervised} (the model we use for intermediate training).
We ensure all entities are code-switched to the same target language in a single sentence, avoiding noise from including too many languages. 
To control the size of the corpus, we generate up to five \textsc{EntityCS} sentences for each English sentence.
In particular, if fewer than five languages have translations available for all the entities in a sentence, we create \textsc{EntityCS} instances with all of them. Otherwise, we randomly select five target languages from the candidates.
If no candidate languages can be found, we do not code-switch the sentence. Instead, we keep it as part of the English corpus. 
Finally, we surround each entity with entity indicators (\texttt{<e>}, \texttt{</e>}).
The statistics of the \textsc{EntityCS} corpus are summarised in \autoref{tab:cs_stats}.
A histogram of the number of sentences and entities per language is shown in~\autoref{sec:nums_cs_corpus}.

\subsection{Masking Strategies}

\begin{figure}[!t]
    \centering

    \begin{subfigure}{\linewidth}
        \includegraphics[width=\linewidth]{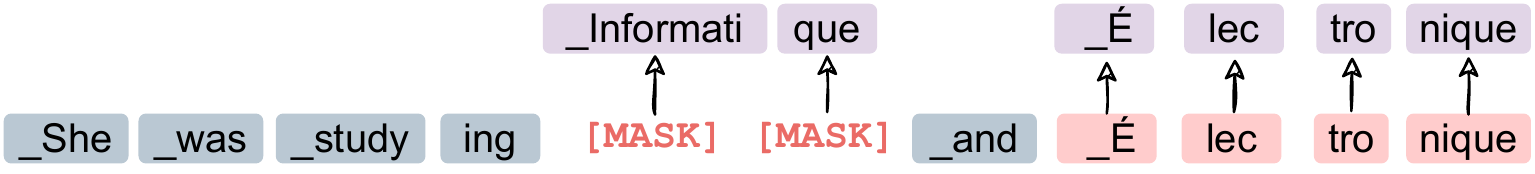}
        \caption{\textbf{W}hole \textbf{E}ntity \textbf{P}rediction (\sc {wep})}
        \label{fig:wep}
         \vspace{1.5ex}
    \end{subfigure}
 
    \begin{subfigure}{\linewidth}
        \includegraphics[width=\linewidth]{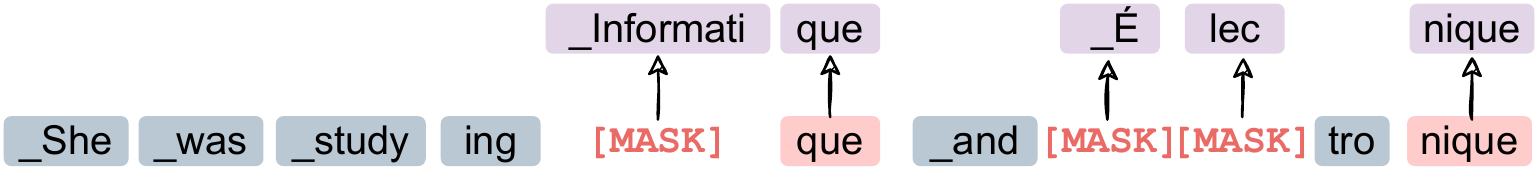}
        \caption{\textbf{P}artial \textbf{E}ntity \textbf{P}rediction (\sc {pep})}
        \label{fig:pep}
          \vspace{1.5ex}
    \end{subfigure}

    \begin{subfigure}{\linewidth}
        \includegraphics[width=\linewidth]{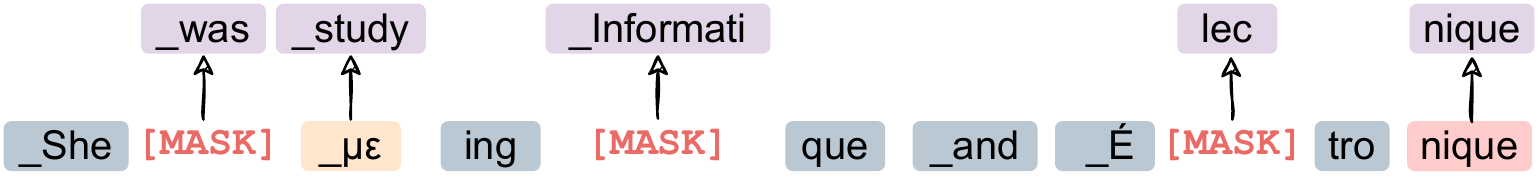}
        \caption{\textbf{P}artial \textbf{E}ntity \textbf{P}rediction with \textbf{MLM} (\sc {pep+mlm})}
        \label{fig:pep_mlm}
    \end{subfigure}
    \caption{Illustration of the proposed masking strategies.
    Random subwords are chosen from the entire vocabulary, and thus can be from different languages. (c) shows a case where ``study'' is replaced with a Greek subword.}
    \label{fig:masking_strategies}
\end{figure}

To test the effectiveness of intermediate training on the generated \textsc{EntityCS} corpus, we experiment with several training objectives using an existing pre-trained language model.
Firstly, we employ the conventional 80-10-10 MLM objective, where 15\% of sentence subwords are considered masking candidates. 
From those, we replace subwords with \texttt{[MASK]} 80\% of the time, with Random subwords (from the entire vocabulary) 10\% of the time, and leave the remaining 10\% unchanged (Same).
To integrate entity-level cross-lingual knowledge into the model, we propose Entity Prediction objectives, where we only mask subwords belonging to an entity.
By predicting the masked entities in \textsc{EntityCS} sentences, we expect the model to capture the semantics of the same entity in different languages.
Two different masking strategies are proposed for predicting entities: Whole Entity Prediction (\textsc{WEP}) and Partial Entity Prediction (\textsc{PEP}).

\begin{table}[!t]
\centering
\scalebox{0.64}{
\begin{tabular}{llcccccccc}
\toprule
\multicolumn{2}{c}{\multirow{2}{*}{\sc \parbox{0.12\textwidth}{Masking Strategy}}} 
& \multicolumn{4}{c}{\sc Entity (\%)}  
& \multicolumn{4}{c}{\sc Non-Entity (\%)}  \\   
&& $p$ 
& \textsc{\textbf{m}ask}
& \textsc{\textbf{r}nd}
& \textsc{\textbf{s}ame}

& $p$ 
& \textsc{\textbf{m}ask} 
& \textsc{\textbf{r}nd} 
& \textsc{\textbf{s}ame} \\  \cmidrule{1-2}    \cmidrule(lr){3-6} \cmidrule(lr){7-10}  

& \sc {mlm} & 15 & 80  & 10  & 10 & 15  & 80 & 10 & 10 \\
\midrule

& \sc {wep} &  100  & 80 & 0 & 20  & 0
&-&-&-
\\
& \sc {pep\textsubscript{mrs}} & 100 & 80 & 10 & 10 &0  
&-&-&-
\\
& \sc {pep\textsubscript{ms}}  & 100 & 80 & 0 & 10 & 0 
&-&-&-
\\
& \sc {pep\textsubscript{m}}   & 100 & 80 & 0 & 0 & 0 
&-&-&-
\\
\midrule

\multirow{4}{*}{\sc {\rotatebox[origin=c]{90}{+mlm}}}
& \sc {wep} & 50& 80 & 0 & 20 &  15  & 80 & 10 & 10 \\
& \sc {pep\textsubscript{mrs}} & 50 & 80 & 10 & 10 &  15  & 80 & 10 & 10 \\
& \sc {pep\textsubscript{ms}} & 50 & 80 & 0 & 10 &  15  & 80 & 10 & 10 \\
& \sc {pep\textsubscript{m}}  & 50 & 80 & 0 & 0 &  15  & 80 & 10 & 10 \\
\bottomrule
\end{tabular}
}
\caption{Summary of the proposed masking strategies. $p$ corresponds to the probability of choosing candidate items (entity/non-entity subwords) for masking.
\textsc{\textbf{m}ask}, \textsc{\textbf{r}nd}, \textsc{\textbf{s}ame} represent the percentage of replacing a candidate with Mask, Random or the Same item.
When combining \textsc{wep/pep} with \textsc{mlm (+mlm)}, we lower $p$ to 50\%.
} 
\label{tab:masking_strategy}
\end{table}

In \textsc{WEP}, motivated by \citet{Sun2019ERNIEER} where whole word masking is also adopted, we consider all the \textit{words} (and consequently subwords) inside an entity as masking candidates. 
Then, 80\% of the time we mask \textit{every} subword inside an entity, and 20\% of the time we keep the subwords intact. 
Note that, as our goal is to predict the entire masked entity, we do not allow replacing with Random subwords, since it can introduce noise and result in the model predicting incorrect entities. 
After entities are masked, we remove the entity indicators \texttt{<e>}, \texttt{</e>} from the sentences before feeding them to the model.
\autoref{fig:wep} shows an example of \textsc{WEP}.

For \textsc{PEP}, we also consider all entities as masking candidates.
In contrast to \textsc{WEP}, we do not force subwords belonging to one entity to be either all masked or all unmasked. 
Instead, each \textit{individual entity subword} is masked 80\% of the time.
For the remaining 20\% of the masking candidates, we experiment with three different replacements. 
First, \textsc{PEP\textsubscript{mrs}}, corresponds to the conventional 80-10-10 masking strategy, where 10\% of the remaining subwords are replaced with Random subwords and the other 10\% are kept unchanged.
In the second setting, \textsc{PEP\textsubscript{ms}}, we remove the 10\% Random subwords substitution, i.e. we predict the 80\% masked subwords and 10\% Same subwords from the masking candidates.
In the third setting, \textsc{PEP\textsubscript{m}}, we further remove the 10\% Same subwords prediction, essentially predicting only the masked subwords. 
An example of \textsc{PEP} is illustrated in \autoref{fig:pep}.

Prior work has proven it is effective to combine Entity Prediction with MLM for cross-lingual transfer~\cite{jiang-etal-2020-x}, therefore we investigate the combination of the Entity Prediction objectives together with MLM on non-entity subwords.
Specifically, when combined with MLM, we lower the entity masking probability ($p$) to 50\% to roughly keep the same overall masking percentage.
\autoref{fig:pep_mlm} illustrates an example of \textsc{PEP} combined with \textsc{MLM} on non-entity subwords.
A summary of the masking strategies is shown in \autoref{tab:masking_strategy}, along with the corresponding masking percentages.

\section{Experimental Setup}

After preparing the \textsc{EntityCS} corpus, we further train an XLM with \textsc{WEP}, \textsc{PEP}, \textsc{MLM} and the joint objectives.
We use the sampling strategy proposed by \citet{XLM}, where high-resource languages are down-sampled and low-resource languages get sampled more frequently.
Since recent studies on pre-trained language encoders have shown that semantic features are highlighted in higher layers~\cite{tenney-etal-2019-bert, rogers-etal-2020-primer}, we only train the embedding layer and the last two layers of the model\footnote{Preliminary experiments where we updated the entire network revealed the model suffered from catastrophic forgetting.} (similarly to \citet{calixto-etal-2021-wikipedia}).
We randomly choose 100 sentences from each language to serve as a validation set, on which we measure the perplexity every 10K training steps.
Details of parameters used for intermediate training can be found in~\autoref{sec:hypers}.

\subsection{Downstream Tasks}
As the \textsc{EntityCS} corpus is constructed with code-switching at the entity level, we expect our models to mostly improve entity-centric tasks.
Thus, we choose the following datasets: WikiAnn~\cite{pan-etal-2017-cross} for NER, X-FACTR~\cite{jiang-etal-2020-x} for Fact Retrieval, MultiATIS++~\cite{xu-etal-2020-end} and MTOP~\cite{li-etal-2021-mtop} for Slot Filling, and XL-WiC \cite{raganato-etal-2020-xl} for WSD\footnote{The result reported on the XL-WiC for prior work is our re-implementation based on \url{https://github.com/pasinit/xlwic-runs}.}. 
More details on the datasets can be found in~\autoref{sec:dataset}.

After intermediate training on the \textsc{EntityCS} corpus, we evaluate the zero-shot cross-lingual transfer of the models on each task by fine-tuning task-specific English training data.
For NER we use the checkpoint with the lowest validation set perplexity during intermediate training.
Similarly, for the probing dataset X-FACTR (only consisting of a test set), we probe models with the lowest perplexity and report the maximum accuracy score for all, single- and multi-token entities between the two proposed decoding methods (independent and confidence-based) from the original paper~\citep{jiang-etal-2020-x}.
For MultiATIS++, MTOP, and XL-WiC datasets, we choose the checkpoints with the best performance on the English validation set\footnote{We observed performance drop for those tasks at later checkpoints.}.
For all experiments, except X-FACTR, we fine-tune models with five random seeds and report average and standard deviation.

\subsection{Pre-Training Languages}
\label{sec:pretraining_langs}
Given the size of the \textsc{EntityCS} corpus, we primarily select a subset from the total 93 languages, that covers most of the languages used in the downstream tasks.
This subset contains 39 languages, from WikiAnn, excluding Yoruba\footnote{Yoruba is not included in the \textsc{EntityCS} corpus, as we only consider languages XLM-R is pre-trained on.}.
We train XLM-R-base~\citep{XLM} on this subset, then fine-tune the new checkpoints on the English training set of each dataset and evaluate all of the available languages.

\section{Main Results}

\begin{table*}[t!]
\centering
\scalebox{0.74}{
\addtolength{\tabcolsep}{-0pt}
\begin{tabular}{ll|lrrcccccl}
\toprule
\multicolumn{2}{c|}{\multirow{3}{*}{\textbf{\textsc{model}}}} 

& {\sc\textbf{ner} \sc(f{\small 1})} 
& \multicolumn{3}{c}{\sc\textbf{fact retr.} (acc.)}
& \multicolumn{4}{c}{\sc\textbf{slot filling} (f{\small 1}, f{\small 1}/acc.)} 
& \multicolumn{1}{c}{\sc\textbf{wsd} (acc.)}   
\\
                       
& & \multicolumn{1}{c}{\sc wikiann} 
& \multicolumn{3}{c}{\sc x-factr}   
& \multicolumn{2}{c}{\sc multiatis++} 
& \multicolumn{2}{c}{\sc mtop} 
& \multicolumn{1}{c}{\sc xl-wic} \\

& & &  \textsc{\textit{all}} & \textsc{\textit{single}} & \textsc{\textit{multi}} & \textsc{\textit{SF}} & \textsc{\textit{SF / Intent}} & \textsc{\textit{SF}} &  \textsc{\textit{SF / Intent}}  & 
\\
\cmidrule(lr){1-2} \cmidrule(lr){3-3} \cmidrule(lr){4-6} \cmidrule(lr){7-8} \cmidrule(lr){9-10} \cmidrule(lr){11-11}

& \textsc{xlm-r\textsuperscript{prior}} 
&  61.8  & 3.5 & 9.4 & 2.6  
&   -- & --
& -- &  -- 
& 58.0
\\

& \textsc{xlm-r\textsuperscript{ours}}
& 61.6~\textsubscript{0.28} 
& 3.5 & 9.4 & 2.6 
& 71.8~\textsubscript{1.96} & 73.0~\textsubscript{0.70} / 89.1~\textsubscript{1.04} 
& \textbf{73.2}~\textsubscript{0.89} & 72.5~\textsubscript{0.78} / 86.0~\textsubscript{0.69} 
& 59.1~\textsubscript{1.52}
\\
\midrule

\multirow{4}{*}{\sc {\rotatebox[origin=c]{90}{\small{\textsc{EntityCS}}}}} 
& \textsc{mlm}
&  63.5~\textsubscript{0.50}
& 2.5 & 6.4 & 1.7
&  72.1~\textsubscript{2.34} & 74.0~\textsubscript{0.69} / 89.6~\textsubscript{1.43}  
&  72.8~\textsubscript{0.60} & 72.7~\textsubscript{0.31} / \textbf{86.3}~\textsubscript{0.41} 
& 59.3~\textsubscript{0.44}
\\

& \textsc{wep}
& 62.4~\textsubscript{0.68} 
& \textbf{6.1} & \textbf{19.4} & 3.0 
& 71.6~\textsubscript{1.20} & 71.7~\textsubscript{0.82} / 89.7~\textsubscript{1.25} 
& 72.2~\textsubscript{0.57} & \textbf{73.0}~\textsubscript{0.47} / 86.0~\textsubscript{0.44} 
& \textbf{60.4}~\textsubscript{0.97} 
\\

& \textsc{pep\textsubscript{ms}}
& 63.3~\textsubscript{0.70} 
& 6.0 & 15.0 & \textbf{4.3} 
&  73.4~\textsubscript{1.70}  & \textbf{74.4}~\textsubscript{0.67} / \textbf{90.0}~\textsubscript{0.90} 
&  71.5~\textsubscript{0.67}  & 72.7~\textsubscript{0.64} / 86.1~\textsubscript{0.51} 
& 60.2~\textsubscript{0.85}
\\

& \textsc{pep\textsubscript{ms}+mlm}  
& \textbf{\textbf{64.4}}~\textsubscript{0.50}  
&  5.7 & 13.9 & 3.9 
& \textbf{74.2}~\textsubscript{0.43} & 74.3~\textsubscript{0.82} / 89.0~\textsubscript{0.87} 
& 73.0~\textsubscript{0.33} & 72.5~\textsubscript{0.57} / 85.8~\textsubscript{0.77} 
&  59.8~\textsubscript{0.75}
\\ 

\bottomrule
\end{tabular}
}
\caption{Average performance across languages on the test set of downstream tasks.
\textsc{xlm-r\textsuperscript{prior}} corresponds to previous reported results with XLM-R-base, referring to \citet{chi-etal-2021-improving} for WikiAnn, \citet{jiang-etal-2020-x} for X-FACTR and \citet{raganato-etal-2020-xl} for XL-WiC. 
\textsc{xlm-r\textsuperscript{ours}} shows our re-implemented results with XLM-R-base.
Results (excluding X-FACTR) are averaged across five seeds with standard deviation reported as a subscript.
}
\label{tab:main_res}
\end{table*}

\begin{table*}[!ht]
\centering
\scalebox{0.72}{
\addtolength{\tabcolsep}{-1.8pt}
\begin{tabular}{l|cccccccccccccccccccc}
\toprule
\sc\textbf{model} & \sc{ar} & \sc{he} & \sc{vi} & \sc{id} & \sc{jv} & \sc{ms} & \sc{tl} & \sc{eu} & \sc{ml} & \sc{ta} & \sc{te} & \sc{af} & \sc{nl} & \sc{en} & \sc{de} & \sc{el} & \sc{bn} & \sc{hi} & \sc{mr} & \sc{ur}   \\ 
\midrule
\sc {xlm-r\textsuperscript{ours}} & 44.6 & 51.9 & 68.3 & 48.6 & 59.6 & 63.3 & 72.5 & \textbf{61.2} & 63.2 & 54.3 & 49.3 & 76.3 & 80.7 & 83.4 & 75.4 & 74.2 & 67.9 & 68.3 & 61.8 & 55.8  \\
\midrule
\sc {pep\textsubscript{ms}}  & 49.6 & 53.0 & 70.0 & 58.5 & \textbf{62.0} & 64.9 & \textbf{75.7} & 59.8 & 63.3 & \textbf{57.7} & 52.1 & 76.4 & 80.9 & 83.8 & 75.1 & 76.3 & 72.5 & 70.1 & \textbf{66.8} & 61.5  \\
\sc {pep\textsubscript{ms}+mlm} & \textbf{51.5} & \textbf{54.0} & \textbf{70.9} & \textbf{61.1} & 59.3 & \textbf{69.9} & 74.6 & 59.3 & \textbf{66.3} & 57.6 & \textbf{54.8} & \textbf{77.9} & \textbf{81.5} & \textbf{84.2} & \textbf{75.5} & \textbf{77.1} & \textbf{74.6} & \textbf{70.7} & 66.3 & \textbf{65.9} \\

\midrule
& \sc{fa} & \sc{fr} & \sc{it} & \sc{pt} & \sc{es} & \sc{bg} & \sc{ru} & \sc{ja} & \sc{ka} & \sc{ko} & \sc{th} & \sc{sw} & \sc{yo} & \sc{my} & \sc{zh} & \sc{kk} & \sc{tr} & \sc{et} & \sc{fi} & \sc{hu} \\ 
\midrule
\sc {xlm-r\textsuperscript{ours}} & 47.6 & 78.0 & 78.2 & 78.9 & 76.2 & 77.3 & 63.9 & 22.9 & 66.4 & 48.8 & 4.3 & \textbf{68.3} & 45.4 & 52.7 & 27.7 & 44.2 & 76.9 & 72.4 & 75.6 & 76.9 \\ 
\midrule
\sc {pep\textsubscript{ms}} & \textbf{55.6} & 78.8 & 78.5 & 78.6 & 75.8 & 78.0 & 66.4 & 21.3 & 67.0 & 50.2 & \textbf{4.6} & 66.9 & 44.7 & 55.2 & 26.9 & 48.9 & 77.4 & \textbf{73.4} & 76.6 & 77.8    \\

\sc {pep\textsubscript{ms}+mlm} 
& 54.2 & \textbf{79.5} & \textbf{78.9} & \textbf{80.1} & \textbf{78.2} & \textbf{79.6} & \textbf{67.7} & \textbf{23.2} & \textbf{68.2} & \textbf{52.1} & 4.0 & 66.4 & \textbf{48.4} & \textbf{56.1} & \textbf{29.8} & \textbf{52.0} & \textbf{78.6} & 71.9 & \textbf{76.8} & \textbf{78.8} \\

\bottomrule
\end{tabular}
}
\caption{F1-score per language on the WikiAnn test set. 
Results are averaged across five seeds.
}
 \label{tab:ner_all_langs}
\end{table*}

\begin{figure*}[!t]
    \centering
    \includegraphics[width=0.95\linewidth]{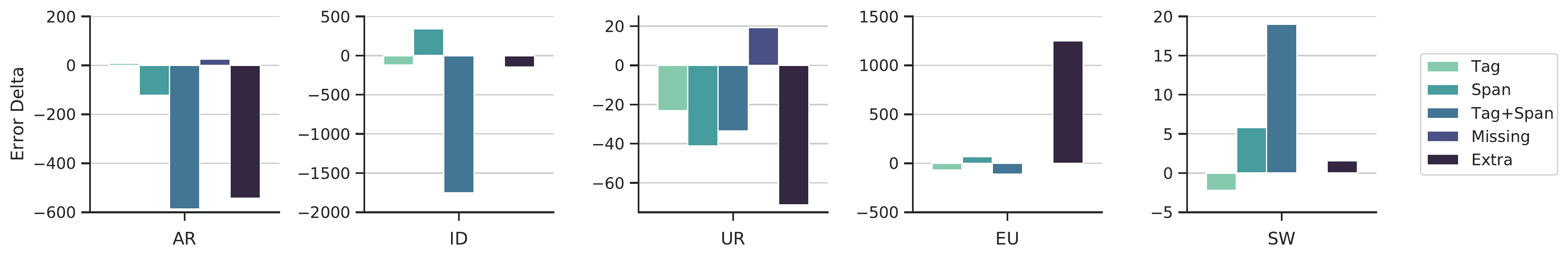}
    \caption{Error Delta (lower is better) for different types of errors in the WikiAnn test set between vanilla XLM-R-base and \textsc{pep\textsubscript{ms}+mlm}.
We show error count differences for AR, ID and UR, the three languages with the largest F1-score improvement, as well as EU and SW, the two languages that underperform the baseline. 
}
\label{fig:error_types}
\end{figure*}

Results are reported in~\autoref{tab:main_res} where we compare models trained on the \textsc{EntityCS} corpus with \textsc{MLM}, \textsc{WEP}, \textsc{PEP\textsubscript{ms}} and \textsc{PEP\textsubscript{ms}+MLM} masking strategies.
For MultiATIS++ and MTOP, we report results of training only Slot Filling (SF), as well as joint training of Slot Filling and Intent Classification (SF/Intent).

\paragraph{Named Entity Recognition} 
For NER, we can see that all models with CS intermediate training show consistent improvement on WikiAnn over the baseline, with \textsc{PEP\textsubscript{ms}+MLM} having +2.8\% absolute improvement. 
This also outperforms XLM-Align~\citep{chi-etal-2021-improving} with 63.7 F1-score, which uses a large amount of parallel data (more results can be found in \autoref{sec:additional_results}).
The classic MLM objective seems equally effective with \textsc{PEP}, possibly because parts of entities are chosen as masking candidates in both. \textsc{WEP} on the other hand, results in lower performance, indicating that full entity prediction from the remaining context is more challenging.
We report performance per language for \textsc{PEP\textsubscript{ms}} and \textsc{PEP\textsubscript{ms}+MLM} in~\autoref{tab:ner_all_langs}\footnote{Per-language results on WikiAnn for other models are reported in \autoref{tab:ner_all_langs_others} in \autoref{sec:additional_results}.}.
We can see that almost all languages benefit from training on the \textsc{EntityCS} corpus.
In our best setting \textsc{PEP\textsubscript{ms}+MLM}, AR, ID, and UR show the largest improvement (around +10\% in AR and ID).
EU and SW on the other hand, result in worse performance compared to the baseline.

As such, we take \textsc{PEP\textsubscript{ms}+MLM}, and further analyse typical NER errors including five categories: Tag, Span, Tag+Span, Missing Extraction and Extra Extraction. 
Missing Extraction occurs when the prediction fails to identify an entity, while Extra Extraction represents errors when a non-entity is wrongly predicted as an entity.
We select EU and SW (lower F1-score than the baseline)\footnote{We do not consider Thai due to its erroneous tokenization in WikiAnn where everything is tokenized into individual \textit{characters}.}, AR, ID, and UR (languages with the largest improvement), and show a delta bar plot in~\autoref{fig:error_types}.
Compared with the baseline, AR, ID, and UR improve consistently Tag+Span and Extra Extraction. 
All but ID, improve Span detection while most languages perform worse in Missing Extraction.
EU and SW on the other hand, result in slightly worse Span and Extra Extraction errors.

We further investigate the reasons for this behaviour. 
In ID, we observe that around 80\% of the Span errors are due to additionally identifying the token \textit{``ALIH''} (means ``moving'', ``changing'' in English) as the start of an entity.
For example, for the input [\textit{``ALIH''}, \textit{``Indofood''}, \textit{``Sukses''}, \textit{``Makmur''}], the gold entity is \textit{``ORG: Indofood Sukses Makmur''}, whereas the model predicts \textit{``ORG: ALIH Indofood Sukses Makmur''}.
This also occurs in XLM-R as 68\% of the span errors.
Consultation with a native speaker reveals this may be an inaccuracy of the dataset (\textit{``ALIH''} should not appear before the actual entities in WikiAnn).
As for EU, the lower overlap between WikiAnn entities and Wikipedia (only 47\% vs 57\% on average across all WikiAnn languages),
might explain the prediction of additional entities not contained in the dataset.

\paragraph{Fact Retrieval}
For X-FACTR, all models trained with Entity Prediction outperform the baseline, whereas \textsc{MLM} is worse than vanilla XLM-R as expected.
For single-token classification, \textsc{WEP} achieves the best results with +10\% gain over XLM-R.
On the other hand in \textsc{PEP\textsubscript{ms}}, we mask part of the entity subwords for prediction, therefore it is best when predicting multi-token entities.
Notably, models trained on large parallel data such as InfoXLM~\cite{chi-etal-2021-infoxlm} and XLM-Align~\cite{chi-etal-2021-improving}, perform poorly with 3\% and 5\% single-token accuracy, respectively, and $<$1\% multi-token accuracy as they focus on alignment at the sentence-level (see \autoref{sec:additional_results}).
Results per language for X-FACTR are available in~\autoref{tab:xfactr_results} of~\autoref{sec:additional_results}.

\paragraph{Slot Filling}
In SF-only training, the best-performing model is 
\textsc{PEP\textsubscript{ms}+MLM}, where we achieve +2.4\% gain over XLM-R on MultiATIS++, also competitive with the best result from XLM-Align (74.4, see \autoref{sec:additional_results}).
In contrast, no improvements can be observed in MTOP over the baseline.
A manual inspection of the dataset reveals that this discrepancy can be attributed to domain differences. 
MultiATIS++ contains entities such as city names, whereas MTOP consists of dialogues with a personal assistant, e.g. setting up reminders, thus fewer entities occur and limit the benefits of entity-centric CS training.
When jointly optimising SF and Intent Classification, models also improve over the baseline on SF (+1.4\% for MultiATIS++ and +0.5\% for MTOP), however with lower gains.
We speculate that the additional intent labels offer complementary information to the task, minimising the impact of external information.

\begin{table}[!t]
\centering
\scalebox{0.65}{
\addtolength{\tabcolsep}{-2.3pt}
\begin{tabular}{l|ccccc|c||ccc|c}
\toprule
\multirow{2}{*}{\sc\textbf{model}}   

& \multicolumn{6}{c||}{\sc {Latin Script}} 
& \multicolumn{4}{c}{\sc {Non Latin Script}} 
  \\
  \cmidrule(lr){2-7} \cmidrule(lr){8-11} 
                
  & \textsc{es} & \textsc{de} 
& \textsc{fr}  & \textsc{pt} & \textsc{tr}  &  \textsc{\textit{avg}}  
& \textsc{zh}   & \textsc{ja} 
& \textsc{hi}
& \textsc{\textit{avg}}    
\\
\cmidrule{1-1} \cmidrule(lr){2-7} \cmidrule(lr){8-11}

\sc {xlm-r\textsuperscript{ours}}
 & {81.5} & 79.8 
& 74.8 & {76.5} & 43.0   & 71.1
& 77.2 & 56.8 
& 50.6 
& 61.5
\\
\midrule
\sc {mlm} 
 & 78.8 & 78.0 & 74.4 & 74.6 & 39.7 & 69.1 & 76.4 & 70.3 & 61.5 & 69.4

\\
\sc {pep\textsubscript{ms}}
 & 79.3 & 79.7 & 75.3 & 76.2 & 45.3 & 71.1 &77.8 & 69.0 & 62.9 & 69.9
 \\
\sc {pep\textsubscript{ms}+mlm} 
& 81.3 & 81.4 & 78.2 & 76.1 & 42.1 & \textbf{71.8} & 78.8 & 68.8 & 65.8 & \textbf{71.1}
\\
\bottomrule
\end{tabular}}
\caption{F1-score (average across five seeds) for languages with Latin and Non-Latin script on MultiATIS++ test set when using SF-only training.}
\label{tab:multiatis_lang}
\end{table}

We then categorise languages in MultiATIS++ based on whether they have the same script as English (Latin) and investigate their performance on SF-only training.
From~\autoref{tab:multiatis_lang} we can see that the models trained on the \textsc{EntityCS} corpus demonstrate notable improvements in languages with non-Latin scripts, +9.6\% on average. 
This indicates that with entity-focused training, models capture information that is especially useful for languages with different scripts than English. More results on MultiATIS++ can be found in~\autoref{tab:multiatis_main} of \autoref{sec:additional_results}.

\paragraph{Word Sense Disambiguation}
For XL-WiC we observe less improvement across tasks, where \textsc{WEP} performs best with +1.3\% over the baseline.
This behaviour can be attributed to the nature of the task, as in our entity-based training objectives, disambiguation is assumed already been addressed and treated as implicit external information.
Notably, when testing XLM-Align that uses parallel data, we observe that it does not improve ambiguous word-level semantics across languages (57\% accuracy).

\section{Analysis}
We conduct further analysis on different masking strategies, their impact across languages and training steps when performing intermediate training on the \textsc{EntityCS} corpus.
We primarily focus on WikiAnn, as it contains the largest amount of languages from the datasets we evaluate on.

\subsection{Performance vs Pre-training Languages}

For WikiAnn, X-FACTR and MultiATIS++, we additionally train \textsc{MLM}, \textsc{WEP} and \textsc{PEP\textsubscript{ms}+MLM} by varying the number of languages in the \textsc{EntityCS} corpus. 
We experiment with using English (no code-switching), the subset of 39 languages (as mentioned in \autoref{sec:pretraining_langs}) and all 93 languages.

\begin{table}[t!]
\raggedright
\scalebox{0.7}{
\addtolength{\tabcolsep}{-0.5pt}
\begin{tabular}{ll|ccrcrr}
\toprule
\multicolumn{2}{l|}{\multirow{2}{*}{\textbf{\textsc{model}}}} &\multirow{2}{*}{\textsc{wikiann}} 
 & \multicolumn{3}{c}{\textsc{x-factr}}&
\multicolumn{2}{c}{\textsc{multiatis++}} \\ 
   \cmidrule(lr){4-6}  \cmidrule(lr){7-8}
\multicolumn{2}{l|}{}         &          &  \textsc{\textit{all}} & \textsc{\textit{single}} & \textsc{\textit{multi}}    &  \textsc{\textit{SF}} &  \textsc{\textit{SF / Intent}}  \\ \midrule
\multicolumn{2}{l|}{\sc {xlm-r\textsuperscript{ours}}}
& 61.6 & 3.5 & 9.4 & 2.6
& 70.6
& 73.0 / 88.9

\\ \midrule
\multirow{3}{*}{\textsc{mlm}}            
& \textsc{en}        
&    61.0
&   1.1  & 2.7  & 0.7     
&  71.5 &   72.1 / 89.6    \\

& \textsc{39}
&   \textbf{63.5}   
& 2.6 & 6.4 &1.7       
& 72.5 & \textbf{73.8} / \textbf{90.2}        \\
 
& \textsc{93}  
&  63.3    
&  \textbf{2.7} & \textbf{6.8}  &  \textbf{1.8} 
& \textbf{72.7}     &    73.4 / 89.6      
\\ \midrule

\multirow{3}{*}{\textsc{wep}}            
& \textsc{en}        
& 61.9   &     
3.3 & 8.5 & 1.6    
&    \textbf{71.8}   &    72.2 / \textbf{91.1}     \\

& \textsc{39}  &  \textbf{62.4}   
&  \textbf{6.1} & \textbf{19.4} &  \textbf{3.0}      
&    71.1  & 71.7 / 89.7       \\
 
& \textsc{93}   &  59.4    
&  5.8 & 18.6 & 2.7
&  70.4          
&    \textbf{72.9} / 90.3    \\ \midrule

\multirow{3}{*}{\sc \parbox{0.05\textwidth}{\textsc{pep\textsubscript{ms}
+mlm}}}
& \textsc{en}      
&     61.2 & 
2.7 & 6.6 & 1.6   
&  71.3       
&    72.3 / 90.7    \\

& \textsc{39}      
&   \textbf{64.4} &  
\textbf{5.7}   &   \textbf{13.9}  &     \textbf{3.9}
&  \textbf{73.4}      
&   \textbf{74.4} / 90.0       \\
 
& \textsc{93}      
&     63.6
&  5.5  &   13.2  &  3.8   
&   72.8   &   72.7 / \textbf{90.8}      \\ 
\bottomrule
\end{tabular}
}
\caption{Results (average over five seeds) with a different number of pre-training languages.}
 \label{tab:pre-training_langs}
\end{table}

From~\autoref{tab:pre-training_langs}, we see that as expected, using English sentences only does not improve the average performance across languages (only Intent Classification accuracy and WikiAnn with \textsc{WEP} increase over the baseline).
However, models trained on English only can benefit English performance, e.g. with an average of +23.1\% gain on single- and +5.6\% on multi-token predictions in X-FACTR over baseline XLM-R using \textsc{WEP}.
When trained with all 93 languages, models with all masking strategies show positive performance over XLM-R, but the majority of results show that training on fewer languages has overall the best performance.
This indicates that including a wider selection of languages does not necessarily contribute to better results and scaling to too many languages still remains non-trivial.
However, note that the subset of 39 languages covers most languages in those downstream tasks. As such, in cases where more languages are tested, increasing the number of pre-training languages may result in better performance.

\subsection{Performance Across Training}

\begin{figure}[t!]
\raggedright
    \includegraphics[width=\linewidth]{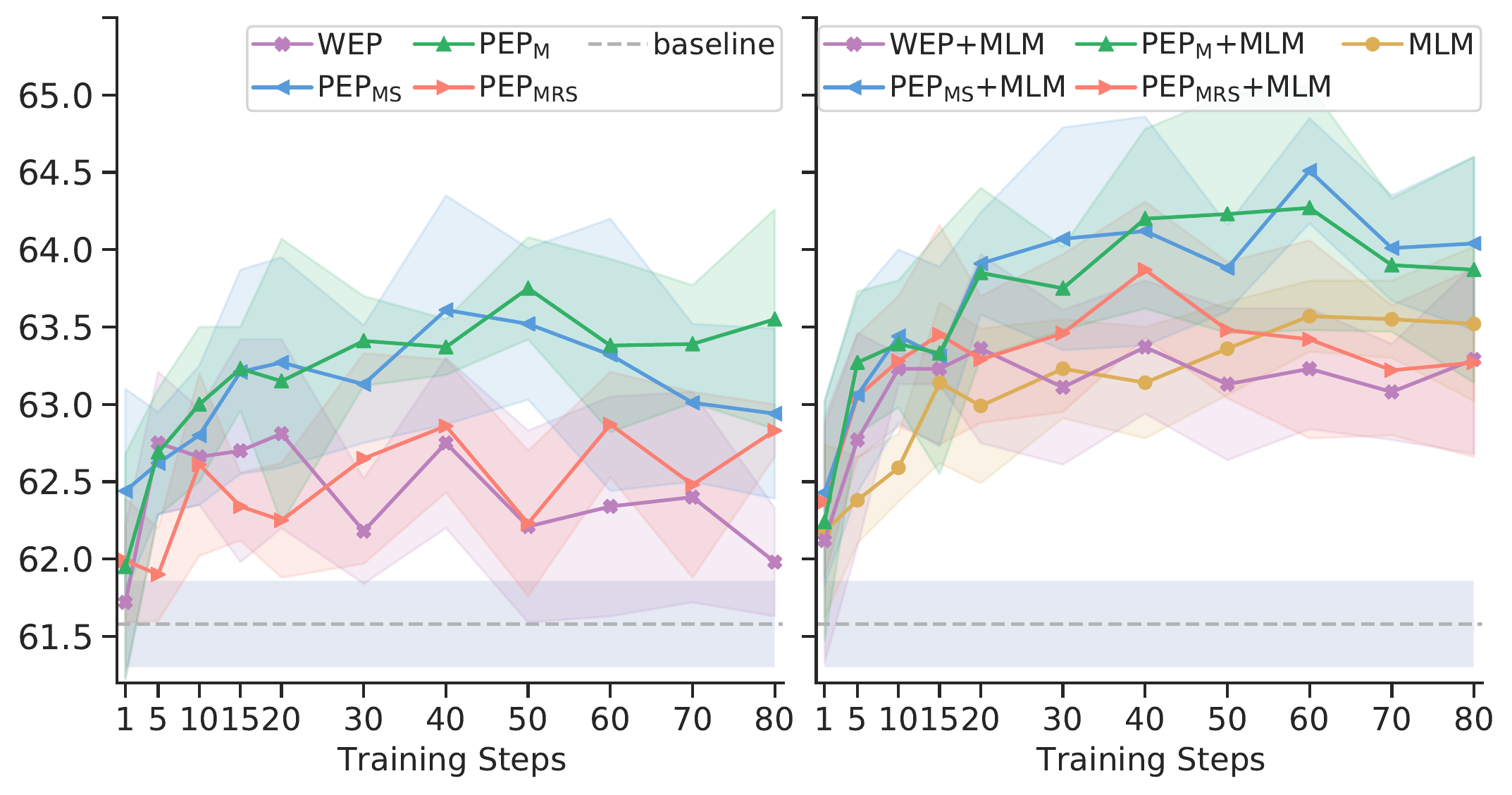}
    \caption{F1-score comparison on WikiAnn test set (average across five seeds) as a function of the number of training steps (in \textit{ten} thousands) with various masking objectives. EP-only strategies are on the left, and EP + MLM strategies are on the right.}
\label{fig:ner39}
\end{figure}

\autoref{fig:ner39} shows a comparison of using different training objectives on the \textsc{EntityCS} corpus, as a function of the number of training steps on the WikiAnn test set.
From the figure, we observe most masking strategies reach a plateau after the middle of training. 
Compared to the objectives that include MLM training, there is a clear increase in performance across the board, further proving that joint training of entities and non-entities is not only beneficial performance-wise but also results in smoother learning curves.
Notably, all objectives surpass the baseline during the entire training.

We can also see that the gains from CS intermediate training do not come from additional training data. 
To prove this, one can look at \autoref{tab:pre-training_langs}, where WikiAnn
F1-score trained on English-only sentences (no CS) shows that extra English-only training data does not improve over the XLM-R baseline (61.6).
This observation can be combined with \autoref{fig:ner39} at step 200K, which corresponds to the steps required for training on the English-only sentences, i.e. the amount of training data that all models see is the same. We can see that all masking strategies achieve an F1 score above 62.3. This shows that the NER performance gain can be attributed to the design of the \textsc{EntityCS} corpus and the training objectives.

\subsection{Random and Same Subword Prediction}

We further investigate the impact of Random subword substitution and the Same subword prediction of the masking candidates when performing \textsc{PEP}.
Comparing \textsc{PEP\textsubscript{mrs}}, \textsc{PEP\textsubscript{ms}} and \textsc{PEP\textsubscript{m}} in~\autoref{fig:ner39},
we can see that including Random subword substitution in \textsc{PEP} results in worse performance, while further removing predicting the Same subwords does not have a significant effect.
Intuitively, Random subword substitution can lead to predicting an incorrect entity and weaken what a model learns, for instance by making the model predict a wrong entity from some remaining (randomly replaced) subwords. 
On the other hand, predicting the Same subword is an easier task that copies the input entity to the output. 
As a result, models seem to neither gain nor lose performance with or without it.
These observations are supported by recent findings \citep{wettig2022should}, although focused on monolingual settings.
We also observe a similar behaviour when combining \textsc{PEP} with \textsc{MLM}.

\subsection{Entity Masking Percentage}

To check the impact of the entity masking candidates percentage during training, we increase the masking probability ($p$) from 50\% to 80\% and 100\% for the best model, \textsc{PEP\textsubscript{ms}+MLM}, and test its effect on the WikiAnn test set.
We observe that further increasing the masking percentage results in a performance drop from 64.4\textsubscript{$\pm{0.50}$} F1-score to 63.8\textsubscript{$\pm{0.67}$} at 80\%, and 64.0\textsubscript{$\pm{0.43}$} at 100\%. 
We speculate that masking too many subwords makes the task of entity prediction from the remaining context more challenging. The close performance between percentages though can be explained by the fact that only two entities exist per sentence on average, as shown in \autoref{tab:cs_stats}.

\section{Related Work}

\paragraph{Cross-Lingual Pre-Training}
Most existing XLMs use parallel data to improve multilingual contextualised word representations for different languages \citep{ouyang-etal-2021-ernie,luo-etal-2021-veco,chi-etal-2021-improving}.
Adapters have been applied to improve zero-shot and few-shot cross-lingual transfer by simply training a handful of model parameters~\cite{pfeiffer-etal-2020-mad, ansell-etal-2021-mad-g}. In addition, meta-learning techniques~\cite{nooralahzadeh-etal-2020-zero, tarunesh-etal-2021-meta} have proven highly effective for fast adaption to new languages~\cite{dou-etal-2019-investigating}.
Compared to those approaches, our method aims to further improve cross-lingual transferability via intermediate training on an entity-based CS corpus created from Wikipedia wikilinks, requiring no parallel data.

\paragraph{Code Switching}
Methods based on code-switching have been successfully applied on cross-lingual model pre-training and fine-tuning on various NLU tasks such as NER~\cite{ner-indian-2020, liu-etal-2021-mulda}, Part-of-Speech Tagging~\cite{ball-garrette-2018-part}, Machine Translation~\cite{srivastava-singh-2020-phinc}, Intent Classification and Slot Filling~\cite{krishnan-etal-2021-multilingual}, as well as on code-switched datasets~\cite{rizal-stymne-2020-evaluating, prasad-etal-2021-effectiveness}.

A major challenge when studying CS is the lack of training data \citep{gupta-etal-2020-semi}.
Existing CS corpora mostly include English and one other language, e.g. English-Chinese, English-Hindi, English-Spanish, extracted from social media platforms~\cite{barik-etal-2019-normalization,xiang-etal-2020-sina, chakravarthi-etal-2020-corpus, ASCEND-corpus}.
Thus, automatic methods for generating CS data in multiple languages have recently been proposed.

\citet{Qin2020CoSDAMLMC} and \citet{conneau-etal-2020-emerging} create CS data from downstream task datasets by randomly switching individual words to a target language using translations from bilingual dictionaries, at the cost of introducing ambiguity errors or switching to words without important content.
\citet{krishnan-etal-2021-multilingual} use CS to improve Intent Classification and Slot Filling.
Rather than switching individual words, they obtain phrase information from the slot labels and generate phrase-level CS sentences via automatic translations. 
\citet{Yang2020AlternatingLM} create CS sentences by randomly substituting source phrases with their target equivalents in parallel sentences after obtaining word alignments.
\citet{jiang-etal-2020-x} select a subset of Wikipedia sentences in four languages that contain multilingual entities from X-FACTR and create CS sentences by switching entities from English to non-English entities \textit{and} vice versa, via Wikidata translations. 
Our work shares several similarities, while the main differences include that we use only the English Wikipedia, create an entity-level English-to-other languages CS corpus including 93 languages in total and test multiple entity prediction training objectives on four downstream tasks.

\paragraph{Knowledge Integration into Language Models}
Large Pre-trained Languages Models (PLM) lack explicit grounding to real-world entities and relations, making it challenging to recover factual knowledge~\cite{Bender2021OnTD}. 
Most current research on knowledge integration into PLMs focuses on monolingual models.
KnowBert~\cite{peters-etal-2019-knowledge} integrates knowledge into BERT~\cite{devlin-etal-2019-bert} by identifying entity spans in the input text and uses an entity linker to retrieve entity embeddings from a KB. 
KEPLER~\cite{KEPLER} jointly optimises MLM and knowledge embeddings with supervision from a KB.
K-Adapter~\cite{wang-etal-2021-k} integrates learnable factual and linguistic knowledge adapters to PLMs by training them in a multi-task setting on relation and dependency-tree prediction.
These models show improvements over the baseline on various tasks including relation classification, entity typing and word sense disambiguation.

Integrating multilingual knowledge into XLMs has also recently been addressed.
\citet{jiang2022xlm} train a model with two knowledge-related tasks, entity prediction and object entailment.
They use WikiData description embeddings in one language (English and non-English) to predict an entity in a target language as a classification task, preserving an entity vocabulary.
\citet{calixto-etal-2021-wikipedia} use Wikipedia articles in 100 languages together with BabelNet~\citep{BabelNet}, a multilingual sense-inventory for WSD, by predicting the WikiData ID of each entity.
Another work taking advantage of entities by \citet{ri-etal-2022-mluke} uses dedicated multi-lingual entity embeddings on 24 languages and outperforms word-based pre-trained models in various cross-lingual transfer tasks.

\section{Conclusions}

In this work, we improve zero-shot transfer on entity-oriented tasks via entity-level code-switching.
We make use of the English Wikipedia and the Wikidata KB to construct an \textsc{EntityCS} corpus by replacing entities in wikilinks with their counterparts in other languages.
We further propose entity-oriented training objectives to improve entity predictions. 
Evaluation of the models on five datasets reveals consistent improvements in NER, Fact Retrieval and WSD over the baseline and prior work that uses large amounts of parallel data, as well as competitive results on Slot Filling.

Interestingly, we found that replacing masking candidates with Random subwords in the conventional masking strategy is harmful for entity prediction, while different masking strategies are optimal for different downstream tasks.
Specifically, Whole Entity Prediction performs best when the emphasis is given to single-token factual knowledge. 
On the other hand, entity typing and multi-token factual retrieval benefit from Partial Entity Prediction, i.e. when the model is given incomplete information on entities and is required to fill in the gaps.
Concurrently jointly predicting non-entity and entity subwords can improve tasks where the entire input context is important for prediction. In addition, our method is particularly beneficial for languages with a non-Latin script.

Our corpus construction process is generic and can be scaled to many more languages, while the proposed masking strategies can be used with any existing language model.
Future work will focus on code-switching beyond entities, such as verbs and phrases, as well as using other sources beyond Wikipedia and Wikidata.

\section*{Limitations}
An important limitation of the work is that before code-switching an entity, its morphological inflection is not checked. This can lead to potential errors as the form of the CS entity might not agree with the surrounding context (e.g. plural). There should be few cases as such, as we are only switching entities. However, this should be improved in a later version of the corpus.
Secondly, the diversity of languages used to construct the \textsc{EntityCS} corpus is restricted to the overlap between the available languages in WikiData and XLM-R pre-training. 
This choice was for a better comparison between models, however, it is possible to extend the corpus with more languages that XLM-R does not cover.

We also acknowledge that the proposed approach was mainly evaluated on entity-centric tasks. Evaluation of more general natural language understanding tasks such as Natural Language Inference is feasible and the impact of the method on them should be explored.
In addition, we only CS from English to other languages, keeping the context in English. This was because fine-tuning is performed on English sentences alone, and we wanted to avoid using raw training sentences from non-English languages (which are naturally much more limited). 
Nevertheless, we believe that CS from non-English articles to English is an important next step, where prior work has shown promising results.
Finally, we did experiments only with base-sized models for speed, though improvements should stand for large models as well.

\section*{Acknowledgements}
We would like to thank Muhammad Idham Habibie for assisting us with the error analysis in Indonesian, as well as the Huawei Noah's Ark London NLP team for providing feedback on an earlier version of the manuscript.
We also acknowledge the MindSpore team for providing technical support\footnote{\url{https://www.mindspore.cn/en}, \\ \url{https://github.com/mindspore-ai}}.

\bibliography{anthology}
\bibliographystyle{acl_natbib}
\appendix

\section{Language Distribution of the \textsc{EntityCS} Corpus}
\label{sec:nums_cs_corpus}

\begin{figure*}
\centering
\includegraphics[width=0.95\textwidth]{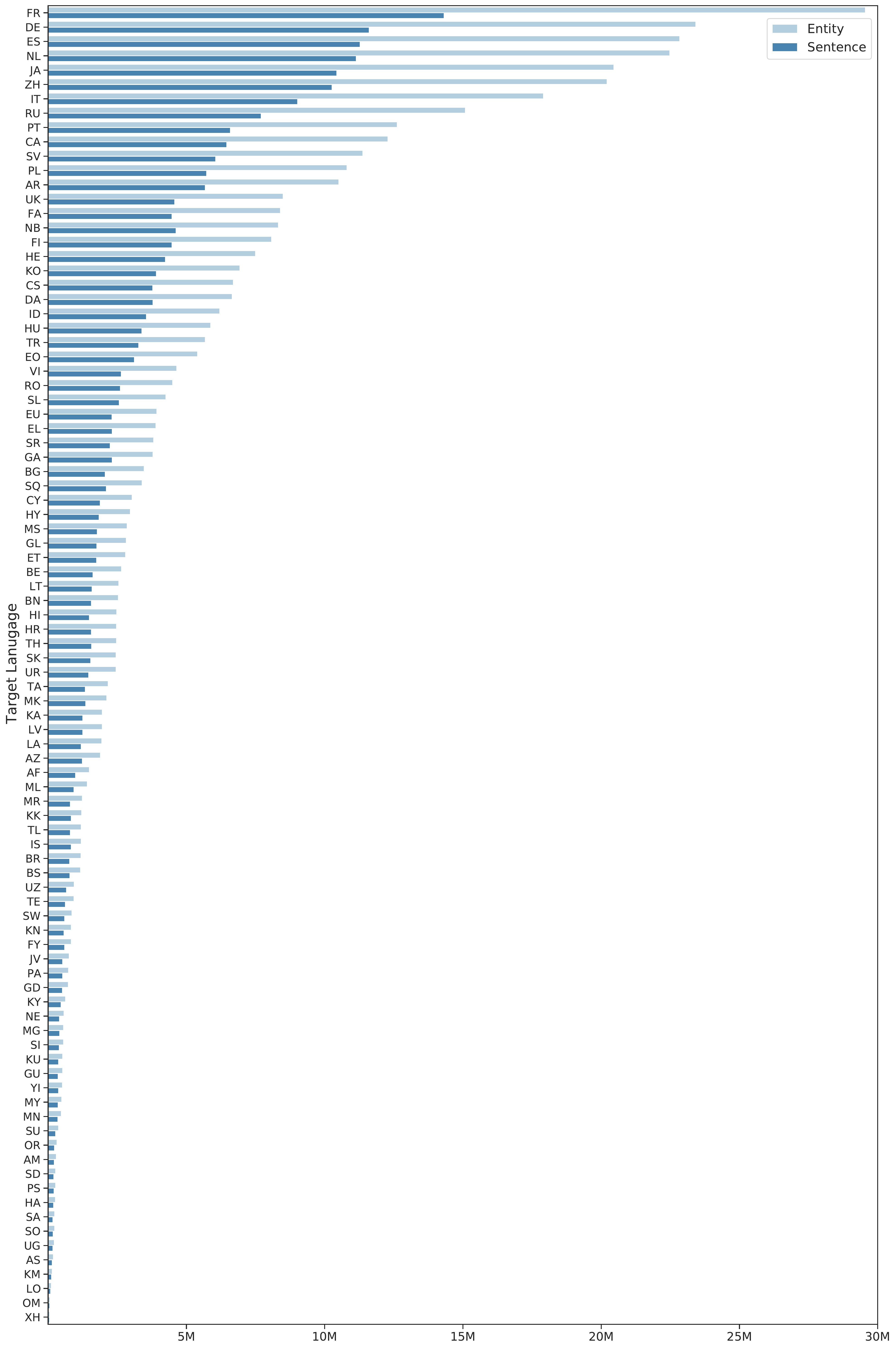}
\caption{Number of Code-Switched Entities and Sentences in the \textsc{EntityCS} corpus.}
\label{fig:corpus_stats}
\end{figure*}

\autoref{fig:corpus_stats} shows the histogram of code-switched entities and sentences in the \textsc{EntityCS} corpus for each languages, except English.

\section{Datasets}
\label{sec:dataset}

We evaluate our models on the following datasets.

\paragraph{WikiAnn}~\citep{pan-etal-2017-cross} is a cross-lingual name tagging and linking dataset based on Wikipedia articles, where named entities are annotated as location (\texttt{LOC}), organisation (\texttt{ORG}) and person (\texttt{PER}) tags following the IOB2 format.
The original dataset contains 282 languages. We evaluate our models on the 40 languages from WikiAnn that are included in the XTREME benchmark~\cite{Hu2020XTREMEAM}.

\paragraph{X-FACTR}~\cite{jiang-etal-2020-x} is a multilingual fact retrieval benchmark similar to LAMA~\cite{petroni-etal-2019-language}.
It probes factual knowledge stored in pre-trained language models by prompt-based fill-in-the-blank cloze queries, covering 23 languages.
X-FACTR includes both single- and multi-token entities, and two decoding methods (independent and confidence-based) are proposed.

\paragraph{MultiATIS++} \cite{xu-etal-2020-end} is an expansion of the Multilingual ATIS~\cite{MATIS} dataset, which includes nine languages (English, Spanish, German, French, Portuguese, Chinese, Japanese, Hindi and Turkish)  from four language families (Indo-European, Sino-Tibetan, Japonic and Altaic).
It contains dialogues in a single domain, Air Travel Information Services.
While processing the dataset, we noticed 14 examples  in the test set do not have a matching number of tokens and slot labels, which we ignore during evaluation.

\paragraph{MTOP}~\cite{li-etal-2021-mtop} is a Multilingual Task-Oriented Parsing dataset that includes six languages from 11 domains that are related to interactions with a personal assistant.
We use the standard flat labels as reported in \citet{li-etal-2021-mtop}.

\paragraph{XL-WiC}~\citep{raganato-etal-2020-xl} is a cross-lingual word disambiguation dataset (Word in Context), formed as a binary classification problem. 
Given a target word and two contexts, the goal is to identify if the word is used in the same sense in both contexts. The dataset contains both nouns and verbs as target words, covers 12 languages and was created as an extension to the English WiC dataset~\citep{pilehvar-camacho-collados-2019-wic}.

\section{Hyper-Parameter Settings}
\label{sec:hypers}


\subsection{Intermediate Training} 

We use 8 NVIDIA V100 32GB GPUs for training our models on the \textsc{EntityCS} corpus, with the Hugging Face library~\citep{wolf-etal-2020-transformers}. 
During fine-tuning, all models were run on a single Nvidia V100 32GB GPU.
We set the batch size to 16 and gradient accumulation steps to 2, resulting in an effective batch size of 256.
For speedup, we employ half-precision (fp16) in the experiments.
In each batch, we allow examples from multiple languages, based on the sampling strategy followed by \citet{XLM}.
We train for a single epoch with a maximum learning rate $5e^{-5}$ and linear decay scheduler, no warmup or weight decay, gradient clipping equal to 1.0, and early stopping if perplexity does not drop after 20 consecutive evaluations (we evaluate every 10K training steps).

\subsection{Downstream Tasks}
For downstream tasks, we evaluate models on the English validation set five times per epoch following \citet{dodge2020fine}.
For fine-tuning XLM-R-base on WikiAnn, MultiATIS++, MTOP and XL-WiC,
we fix the number of training epochs to 10, gradient clipping to $1.0$ and maximum sequence length to 128. We select the batch size from \{$8$, $32$\}, learning rate from \{$1e^{-5}$, $2e^{-5}$, $3e^{-5}$, $4e^{-6}$, $5e^{-6}$, $6e^{-6}$\}, and warm up ratio from \{$0$, $0.1$\}.
The best hyper-parameters per task are reported in~\autoref{tab:best_hypers}.

\begin{table*}[!t]
    \centering
    \scalebox{0.78}{
    \addtolength{\tabcolsep}{0pt}
    \begin{tabular}{l|llllll}
    \toprule
    \multirow{2}{*}{\sc{Parameter}}  & 
    \multirow{2}{*}
    {\textsc{wiki-ann}}
    & \multicolumn{2}{c}{\textsc{multiatis++}}  & \multicolumn{2}{c}{\textsc{mtop}}
    & 
        \multirow{2}{*}
        {\textsc{xl-wic}}
    
    \\ 
        \cmidrule(lr){3-4} \cmidrule(lr){5-6}

    &     & \textit{SF} & \textit{Joint} &\textit{SF} & \textit{Joint} \\
        
        \midrule
        \textsc{Learning Rate}     & $1e^{-5}$ & $3e^{-5}$ & $3e^{-5}$ & $2e^{-5}$ & $3e^{-5}$ &$1e^{-5}$ \\
        \textsc{Warmup Ratio}        &   0.1 & 0.0  &0.0 & 0.1 & 0.1 &0.0 \\
        \textsc{Batch Size}        & 8 & 8 & 8 & 8 & 8 &8 \\
    \bottomrule
    \end{tabular}
    }
    \caption{Best hyper-parameters used for the datasets.}
    \label{tab:best_hypers}
\end{table*}

\section{Additional Results}
\label{sec:additional_results}

We report per-language results in the following tables. 
WikiAnn results can be found in \autoref{tab:ner_all_langs_others}.
X-FACTR results in \autoref{tab:xfactr_results}, MultiATIS++ Slot Filling-only training in \autoref{tab:multiatis_main},
and XL-WiC in \autoref{tab:xlwic_results}.

Comparison with InfoXLM and XLM-Align described in the paper is also summarised in \autoref{tab:xlms-compare}. 
These results are not included in the main table as InfoXLM and XLM-Align use parallel data, therefore the comparison is not fair.

\begin{table*}[t!]
\centering
\scalebox{0.76}{
\addtolength{\tabcolsep}{-0.2pt}
\begin{tabular}{ll|lcrcccc}
\toprule
\multicolumn{2}{c|}{\multirow{3}{*}{\textbf{\textsc{model}}}} 

& {\sc\textbf{ner} \sc(f{\small 1})} 
& \multicolumn{3}{c}{\sc\textbf{fact retr.} (acc.)}
& \multicolumn{2}{c}{\sc\textbf{slot filling} (f{\small 1})} 
& \multicolumn{1}{c}{\sc\textbf{wsd} (acc.)}   
\\
                       
& & \multicolumn{1}{c}{\sc wikiann} 
& \multicolumn{3}{c}{\sc x-factr}   
& \multicolumn{1}{c}{\sc multiatis++} 
& \multicolumn{1}{c}{\sc mtop} 
& \multicolumn{1}{c}{\sc xl-wic} \\

& & &  \textsc{\textit{all}} & \textsc{\textit{single}} & \textsc{\textit{multi}} & \textsc{\textit{SF}} & \textsc{\textit{SF}}  & 
\\
\cmidrule(lr){1-2} \cmidrule(lr){3-3} \cmidrule(lr){4-6} \cmidrule(lr){7-7} \cmidrule(lr){8-8} \cmidrule(lr){9-9}
& \textsc{xlm-r\textsuperscript{ours}}
& 61.6~\textsubscript{0.28} 
& 3.5 & 9.4 & 2.6 
& 70.6~\textsubscript{1.55} 
& 72.3~\textsubscript{0.98}
& 59.1~\textsubscript{1.52}
\\
& \textsc{infoxlm}\small{~\cite{chi-etal-2021-infoxlm}}
& 62.8 &  1.1  & 3.3 &  0.6
&  73.9 \textsubscript{1.95} 
& 74.7 \textsubscript{0.30}   
& 56.9~\textsubscript{0.81} 

\\
& \textsc{xlm-align}\small{~\cite{chi-etal-2021-improving}}
& 63.7 &  1.5  & 5.0 & 1.0
&  \textbf{74.4} \textsubscript{0.29} 
& \textbf{74.9} \textsubscript{0.36}   
& 56.9~\textsubscript{1.22} 
\\

\midrule

\multirow{3}{*}{\sc {\rotatebox[origin=c]{90}{\small{\textsc{EntityCS}}}}} 

& \textsc{wep}
& 62.4~\textsubscript{0.68} 
& \textbf{6.1} & \textbf{19.4} & 3.0 
& 71.6~\textsubscript{1.20}
& 73.2~\textsubscript{0.89}
& \textbf{60.4}~\textsubscript{0.97} 
\\

& \textsc{pep\textsubscript{ms}}
& 63.3~\textsubscript{0.70} 
& 6.0 & 15.0 & \textbf{4.3}
& 73.4~\textsubscript{1.70}  
& 71.5~\textsubscript{0.67}  
& 60.2~\textsubscript{0.85}
\\

& \textsc{pep\textsubscript{ms}+mlm}  
& \textbf{64.4}~\textsubscript{0.50}  
& 5.7 & 13.9 & 3.9 
& 74.2~\textsubscript{0.43}
& 73.0~\textsubscript{0.33} 
& 59.8~\textsubscript{0.75}
\\ 

\bottomrule
\end{tabular}
}
\caption{Comparison with models using parallel data.}
\label{tab:xlms-compare}
\end{table*} 

\begin{table*}[!ht]
\centering
\scalebox{0.72}{
\addtolength{\tabcolsep}{-2pt}
\begin{tabular}{l|cccccccccccccccccccc}
\toprule
\sc\textbf{model} & \sc{ar} & \sc{he} & \sc{vi} & \sc{id} & \sc{jv} & \sc{ms} & \sc{tl} & \sc{eu} & \sc{ml} & \sc{ta} & \sc{te} & \sc{af} & \sc{nl} & \sc{en} & \sc{de} & \sc{el} & \sc{bn} & \sc{hi} & \sc{mr} & \sc{ur}   \\ 
\midrule
\sc {xlm-r\textsuperscript{ours}} & 44.6 & 51.9 & 68.3 & 48.6 & 59.6 & 63.3 & 72.5 & 61.2 & 63.2 & 54.3 & 49.3 & 76.3 & 80.7 & 83.4 & 75.4 & 74.2 & 67.9 & 68.3 & 61.8 & 55.8  \\
\midrule
\sc {mlm}  
& 50.7 & 53.7 & 72.7 & 56.4 & 59.2 & 68.4 & 75.1 & 58.4 & 65.1 & 58.1 & 53.0 & 76.3 & 80.9 & 84.2 & 75.2 & 76.3 & 73.9 & 69.9 & 64.5 & 67.0 \\
\sc {wep} 
& 49.9 & 52.4 & 69.8 & 57.4 & 60.1 & 66.7 & 74.0 & 60.1 & 60.8 & 56.1 & 48.2 & 76.5 & 80.3 & 83.8 & 74.7 & 74.5 & 70.8 & 67.5 & 61.1 & 60.7 \\
\sc {pep\textsubscript{mrs}}  
& 47.1 & 52.6 & 69.8 & 56.0 & 60.1 & 62.4 & 74.8 & 56.1 & 61.6 & 56.1 & 50.9 & 77.9 & 81.4 & 83.8 & 75.4 & 74.8 & 69.6 & 68.3 & 64.1 & 48.7 \\
\sc {pep\textsubscript{m}}  
& 47.7 & 52.9 & 68.9 & 59.1 & 63.1 & 65.5 & 76.3 & 60.0 & 64.0 & 57.5 & 51.6 & 76.8 & 80.9 & 83.9 & 75.1 & 76.5 & 73.0 & 69.6 & 65.8 & 63.3\\
\sc {wep+mlm}  
& 50.3 & 53.2 & 69.8 & 60.8 & 60.7 & 69.8 & 74.5 & 59.2 & 64.8 & 57.2 & 51.7 & 76.4 & 80.9 & 84.1 & 75.4 & 75.2 & 72.1 & 68.9 & 63.9 & 58.6 \\
\sc {pep\textsubscript{mrs}+mlm}  
& 46.7 & 53.6 & 69.6 & 64.0 & 60.2 & 69.2 & 74.3 & 57.5 & 65.6 & 55.8 & 52.3 & 77.5 & 81.3 & 84.1 & 75.5 & 76.2 & 72.2 & 68.4 & 64.8 & 60.1\\
\sc {pep\textsubscript{m}+mlm}  
& 52.4 & 53.5 & 70.4 & 60.3 & 59.7 & 69.2 & 75.5 & 58.4 & 66.6 & 58.1 & 54.5 & 77.7 & 81.2 & 84.1 & 75.3 & 76.2 & 74.2 & 70.0 & 67.1 & 64.5 \\

 \midrule
& \sc{fa} & \sc{fr} & \sc{it} & \sc{pt} & \sc{es} & \sc{bg} & \sc{ru} & \sc{ja} & \sc{ka} & \sc{ko} & \sc{th} & \sc{sw} & \sc{yo} & \sc{my} & \sc{zh} & \sc{kk} & \sc{tr} & \sc{et} & \sc{fi} & \sc{hu} \\ \midrule
\sc {xlm-r\textsuperscript{ours}} & 47.6 & 78.0 & 78.2 & 78.9 & 76.2 & 77.3 & 63.9 & 22.9 & 66.4 & 48.8 & 4.3 & 68.3 & 45.4 & 52.7 & 27.7 & 44.2 & 76.9 & 72.4 & 75.6 & 76.9 \\ 
\midrule
\sc {mlm}  
&51.6 & 79.0 & 78.6 & 79.5 & 77.6 & 78.6 & 67.2 & 22.7 & 66.1 & 50.8 & 2.5 & 65.1 & 42.9 & 55.7 & 29.7 & 50.7 & 77.8 & 71.4 & 76.2 & 78.1 \\ 
\sc {wep} 
& 50.9 & 77.6 & 77.7 & 77.3 & 74.1 & 78.7 & 66.3 & 20.7 & 64.8 & 52.0 & 2.5 & 65.8 & 50.4 & 52.6 & 26.1 & 52.1 & 75.5 & 71.9 & 75.8 & 76.6 \\
\sc {pep\textsubscript{mrs}}
& 53.0 & 78.9 & 78.6 & 78.7 & 77.1 & 78.6 & 67.3 & 21.9 & 63.6 & 51.4 & 3.7 & 66.2 & 45.9 & 54.6 & 26.6 & 49.1 & 78.0 & 72.8 & 77.2 & 77.7 \\
\sc {pep\textsubscript{m}} 
& 55.3 & 78.5 & 78.4 & 78.6 & 74.3 & 78.2 & 67.2 & 21.0 & 66.7 & 50.0 & 5.0 & 66.8 & 52.3 & 56.9 & 26.7 & 48.4 & 77.6 & 73.2 & 76.6 & 77.3 \\
\sc {wep+mlm}  
&53.4 & 78.1 & 78.3 & 79.0 & 74.9 & 78.3 & 66.6 & 23.0 & 65.8 & 50.4 & 2.1 & 63.9 & 44.9 & 56.6 & 29.1 & 51.0 & 75.2 & 71.8 & 76.9 & 77.2\\
\sc {pep\textsubscript{mrs}+mlm}   
& 54.3 & 79.2 & 79.1 & 80.0 & 76.4 & 79.1 & 67.6 & 23.6 & 66.1 & 50.9 & 2.4 & 66.6 & 41.4 & 54.5 & 31.1 & 51.8 & 78.4 & 71.4 & 77.1 & 78.7 \\
\sc {pep\textsubscript{m}+mlm}   
& 50.0 & 79.9 & 78.9 & 79.7 & 78.4 & 79.3 & 68.2 & 22.7 & 67.7 & 51.1 & 3.3 & 64.5 & 43.7 & 56.6 & 29.3 & 51.7 & 78.2 & 72.0 & 76.8 & 78.6\\
\bottomrule
\end{tabular}
}
\caption{F1-score per language on the WikiAnn test set. 
Results are averaged across five seeds.
}
 \label{tab:ner_all_langs_others}
\end{table*}

\begin{table*}[ht]
\centering
\scalebox{0.8}{
\begin{tabular}{l|ccccccccc}
\toprule
\sc\textbf{model}  & \textsc{en}  & \textsc{es} & \textsc{de} & \textsc{hi} & \textsc{fr}  & \textsc{pt}   & \textsc{zh}   & \textsc{ja} & \textsc{tr}       \\
\midrule
\textsc{xlm-r\textsuperscript{ours}} 
& 95.6~\textsubscript{0.15} & {81.5}~\textsubscript{0.71} & 79.8~\textsubscript{2.04} & 50.6~\textsubscript{5.35} & 
74.8~\textsubscript{1.90} & {76.5}~\textsubscript{1.14} & 77.2~\textsubscript{2.06} & 56.8~\textsubscript{4.99} & 43.0~\textsubscript{2.72} 
\\
\midrule
\textsc{mlm}   
& 95.6~\textsubscript{0.16} & 78.8~\textsubscript{2.88} & 78.0~\textsubscript{2.56} & 61.5~\textsubscript{7.26} & 74.4~\textsubscript{3.18} & 74.6~\textsubscript{1.39} & 76.4~\textsubscript{1.81} & 70.3~\textsubscript{2.00}  & 39.7~\textsubscript{4.09}    \\
\textsc{wep}  
& 95.7~\textsubscript{0.15} & 79.9~\textsubscript{1.34} & 80.3~\textsubscript{0.58} & 52.7~\textsubscript{4.15} & 75.6~\textsubscript{0.87}& 76.3~\textsubscript{0.63} & 78.1~\textsubscript{1.43}  & 60.7~\textsubscript{7.07}    & 40.4~\textsubscript{4.40}  \\
\textsc{pep\textsubscript{ms}}    
& 95.3~\textsubscript{0.06} & 79.3~\textsubscript{2.60} & {79.7}~\textsubscript{2.28} & 62.9~\textsubscript{2.30} & 75.3~\textsubscript{2.10} & 76.2~\textsubscript{1.60} & {77.8}~\textsubscript{1.30} & {69.0}~\textsubscript{4.90} & {45.3}~\textsubscript{2.50}
 \\
\textsc{pep\textsubscript{ms}+mlm}  & 
{95.6}~\textsubscript{0.10} & 81.3~\textsubscript{1.90} & 81.4~\textsubscript{0.90} & {65.8}~\textsubscript{2.20} & 78.2~\textsubscript{0.30} & {76.1}~\textsubscript{1.00} & 78.8~\textsubscript{0.60} & 68.8~\textsubscript{3.30}   & 42.1~\textsubscript{3.30}  \\
\bottomrule
\end{tabular}}
\caption{F1-score (average across five seeds) on MultiATIS++ Slot Filling-only training.}
\label{tab:multiatis_main}
\end{table*}


\begin{table*}
    \centering
    \scalebox{0.75}{
        \addtolength{\tabcolsep}{-2pt}
    \begin{tabular}{l|cccccccccccc}
    \toprule
      \sc\textbf{model} &  \textsc{bg} &	\textsc{da}	 & \textsc{et} &	\textsc{fa} &	\textsc{hr}	 & \textsc{ja}  &	\textsc{ko} &	\textsc{nl}	  & \textsc{zh}	  & \textsc{de} &	\textsc{fr}  &	\textsc{it} \\ \midrule
        
        \textsc{xlm-r}\textsuperscript{ours} 
        & 57.5~\textsubscript{1.03}	
        & 60.6~\textsubscript{2.06}
        & 61.7~\textsubscript{3.23}
        & 62.4~\textsubscript{1.05}
        & 61.7~\textsubscript{2.93}
        & 54.0~\textsubscript{1.56}
        & 62.4~\textsubscript{1.99}
        & 61.5~\textsubscript{1.94}
        & 56.4~\textsubscript{3.83}
        & 57.7~\textsubscript{1.58}
        & 56.4~\textsubscript{1.40}
        & 57.1~\textsubscript{1.35} \\ \midrule
        
        \textsc{mlm}
        &  59.3~\textsubscript{0.85}
        &  59.0~\textsubscript{0.72}	
        &  60.6~\textsubscript{1.15}	
        &  63.5~\textsubscript{1.41}	
        &  62.3~\textsubscript{1.87}	
        &  52.6~\textsubscript{1.26}	
        &  63.1~\textsubscript{1.38}	
        &  62.3~\textsubscript{0.76}	
        &  52.4~\textsubscript{1.03}	
        &  57.2~\textsubscript{0.54}	
        &  56.6~\textsubscript{0.33}	
        &  58.0~\textsubscript{1.09}
        \\
        												
        \textsc{wep}
        & 59.0~\textsubscript{1.84}	
        & 61.3~\textsubscript{1.13}	
        & 62.2~\textsubscript{0.74}	
        & 64.9~\textsubscript{1.06}	
        & 63.7~\textsubscript{2.40}	
        & 54.7~\textsubscript{2.69}	
        & 64.6~\textsubscript{0.74}	
        & 63.8~\textsubscript{0.77}	
        & 55.2~\textsubscript{3.30}	
        & 59.6~\textsubscript{1.04}	
        & 57.0~\textsubscript{0.98}
        & 59.1~\textsubscript{1.03} 
        \\    
        
        \textsc{pep\textsubscript{ms}}
		& 59.4~\textsubscript{0.93}	
		& 60.7~\textsubscript{1.03}
		& 64.4~\textsubscript{1.72}	
		& 63.5~\textsubscript{1.51}	
		& 64.2~\textsubscript{1.85}	
		& 53.6~\textsubscript{2.60}	
		& 64.6~\textsubscript{2.88}	
		& 63.9~\textsubscript{0.79}	
		& 52.8~\textsubscript{3.01}	
		& 59.5~\textsubscript{1.79}	
		& 57.4~\textsubscript{0.64}	
		& 58.8~\textsubscript{1.25}
		\\					
												
        \textsc{pep\textsubscript{ms}+mlm}
        & 59.7~\textsubscript{1.04}
        & 60.9~\textsubscript{1.20}	
        & 63.9~\textsubscript{1.02}	
        & 63.1~\textsubscript{1.63}	
        & 63.8~\textsubscript{1.89}	
        & 53.2~\textsubscript{2.18}	
        & 62.1~\textsubscript{3.10}	
        & 63.0~\textsubscript{1.01}	
        & 53.2~\textsubscript{2.24}	
        & 59.0~\textsubscript{0.90}	
        & 57.3~\textsubscript{0.58}	
        & 58.3~\textsubscript{1.00}
        \\
    \bottomrule
    \end{tabular}
    }
    \caption{XL-WiC test set accuracy (average across five seeds) across languages.}
    \label{tab:xlwic_results}
\end{table*}


\begin{table*}[t!]
    \centering
    \addtolength{\tabcolsep}{-1.5pt}
    \scalebox{0.65}{
    \begin{tabular}{p{0.3cm}rrrrrrrrrrrrrrrrrrrrrrrrrr}
    \toprule
     \multicolumn{3}{l}{\sc{model}} & &
    \textsc{avg} & \textsc{en} & \textsc{fr} & \textsc{nl} & \textsc{es} &	\textsc{ru} & \textsc{zh} & \textsc{he} & \textsc{tr} &	\textsc{ko} & \textsc{vi} & \textsc{el} & \textsc{mr} &	\textsc{ja} & \textsc{hu} & \textsc{bn} & \textsc{ceb} & \textsc{war} & \textsc{tl} & \textsc{sw} & \textsc{pa} & \textsc{mg} & \textsc{ilo}
    \\ 
    \midrule
    \multirow{6}{*}{\rotatebox[origin=c]{90}{\textsc{xlm-r}\textsuperscript{\textsc{ours}}}} & \multirow{6}{*}
    & \multirow{3}{*}{\rotatebox[origin=c]{90}{\textsc{ind}}} 
    &	\sc{A}
    & 3.5 &	8.2 &	4.7 &	4.4 &	6.5 &	5.3 &	4.6 &	2.5 &	3.1 &	5.1 &	8.5 &	6.3 &	2.7 &	2.3 &	0.9 &	0.1 &	1.4 &	1.2 &	2.8	 & 3.7 &	0.2 &	1.9 &	0.1  \\
     
    && & \sc{S}
    & 9.4  &	15.2 &	11.3 &	11.0 &	13.4 &	14.4 &	11.9 &	12.3 &	4.0 &	16.7 &	14.2 &	27.3 &	19.5 &	9.2 &	2.2 &	0.0 &	1.7 &	1.3 &	5.1 &	5.6 &	5.8 &	3.7 &	0.4  \\
    
        && & \sc{M}
    & 2.1 &	3.3 &	2.3 &	2.6 &	3.3 &	3.8 &	4.5 &	2.2 &	2.5 &	2.6 &	5.1 &	2.9 &	1.1 &	2.1 &	0.2 &	0.1 &	1.0 &	1.1 &	1.4 &	1.9 &	0.0 &	1.6 &	0.0  \\ \cmidrule{3-27}
    
    && \multirow{3}{*}{\rotatebox[origin=c]{90}{\textsc{conf}}} 
    & 	\sc{A}
    & 3.3  &	4.4 &	2.9	 & 2.7  &	4.3 &	5.5 &	5.3	 & 3.0 &	3.0 &	5.6 &	9.5 &	7.3 &	3.4 &	4.4 &	0.9 &	0.1 &	1.2 &	1.1 &	2.3 &	2.9 &	0.6 &	1.8 &	0.5  \\
    
        && & \sc{S}
    & 7.5&	5.2 &	4.4 &	3.6 &	4.9 &	14.2 &	11.8 &	11.4 &	3.9 &	15.9 &	12.6 &	25.6 &	18.9 &	8.8 &	2.0 &	0.0 &	1.4 &	1.4 &	4.4 &	4.3 &	5.8 &	3.5 &	0.5  \\
    
        && & \sc{M}
    & 2.6 &	3.9 &	2.3 &	2.7	 & 4.2 &	4.1 &	5.2 &	2.7 &	2.4 &	3.4 &	7.0 &	4.3 &	2.07 &	4.2 &	0.3 &	0.1 &	1.0 &	1.1 &	1.3 &	1.9 &	0.4 &	1.5&	0.5 \\ \midrule

    \multirow{6}{*}{\rotatebox[origin=c]{90}{\textsc{mlm}}}  & \multirow{6}{*}{\sc 39}
    & \multirow{3}{*}{\rotatebox[origin=c]{90}{\textsc{ind}}} 
       & 	\sc{A}	 
    & 2.3 & 2.1 & 3.7 & 2.9 & 3.9 & 2.9 & 1.9 & 3.4 & 1.2 & 5.0 & 4.6 & 4.2 & 3.6 & 0.3 & 0.7 & 0.0 & 2.1 & 1.0 & 1.4 & 5.2 & 0.0 & 0.0 & 0.1 \\
	    && & \sc{S}	 
	& 6.4 & 5.1 & 8.7 & 6.4 & 9.4 & 6.0 & 8.3 & 8.7 & 3.1 & 16.6 & 9.1 & 19.3 & 17.9 & 2.5 & 1.8 & 0.6 & 2.9 & 1.1 & 4.4 & 8.3 & 0.3 & 0.5 & 0.1 \\
	    && & \sc{M} 
	& 1.3 & 0.9 & 2.0 & 2.0 & 1.9 & 2.0 & 1.8 & 1.9 & 0.6 & 2.3 & 2.4 & 2.8 & 2.1 & 0.2 & 0.4 & 0.0 & 1.8 & 1.0 & 0.5 & 2.2 & 0.0 & 0.0 & 0.1 \\ \cmidrule{3-27}

    && \multirow{3}{*}{\rotatebox[origin=c]{90}{\textsc{conf}}} 
         & 	\sc{A}	 &2.5 & 2.5 & 3.6 & 2.9 & 4.3 & 2.6 & 2.0 & 4.8 & 1.1 & 5.7 & 6.3 & 5.2 & 4.2 & 0.4 & 0.6 & 0.1 & 2.0 & 1.0 & 1.2 & 5.2 & 0.0 & 0.0 & 0.1\\
	&& & \sc{S} & 5.9 & 4.9 & 7.6 & 5.9 & 9.0 & 4.4 & 7.6 & 7.4 & 2.5 & 16.1 & 8.5 & 17.2 & 16.7 & 2.5 & 1.6 & 0.6 & 2.8 & 1.1 & 3.9 & 7.8 & 0.3 & 0.5 & 0.0  \\
	&& & \sc{M} & 	1.7 & 1.8 & 2.3 & 2.2 & 2.6 & 2.3 & 1.9 & 3.4 & 0.5 & 3.4 & 4.6 & 4.2 & 2.9 & 0.3 & 0.4 & 0.0 & 1.7 & 1.0 & 0.4 & 2.4 & 0.0 & 0.0 & 0.1\\ \midrule
         
    \multirow{18}{*}{\rotatebox[origin=c]{90}{\textsc{wep}}}  & \multirow{6}{*}{\sc en}
    & \multirow{3}{*}{\rotatebox[origin=c]{90}{\textsc{ind}}} 
    & \textsc{A}
    & 3.3 & 18.2 & 6.1 & 6.0 & 5.8 & 1.1 & 0.4 & 0.4 & 1.1 & 0.5 & 8.0 & 3.5 & 0.4 & 0.6 & 3.7 & 0.0 & 3.5 &   0.6 
    & 5.0 & 4.2 & 0.1 & 1.7 & 1.6  \\ 
    
        && & \textsc{S}
    & 8.5 & 38.3 & 16.4 & 18.7 & 14.9 & 4.4 & 3.4 & 1.4 & 5.6 & 2.7 & 16.8 & 7.3 & 4.1 & 2.5 & 8.5 & 0.0 & 6.9 & 2.6 & 9.7 & 10.3 & 0.0 & 6.9 & 5.4\\ 
    
        && & \textsc{M}
    & 1.6 & 9.4 & 2.7 & 2.9 & 2.9 & 0.6 & 0.3 & 0.4 & 0.3 & 0.3 & 3.5 & 1.2 & 0.1 & 0.5 & 2.6 & 0.0 & 1.5 & 0.3 & 1.5 & 1.9 & 0.1 & 1.1 & 0.5
    \\ \cmidrule{3-27}
    
    && \multirow{3}{*}{\rotatebox[origin=c]{90}{\textsc{conf}}} 
       &  \textsc{A}
    & 3.1 & 16.2 & 6.4 & 5.6 & 5.4 & 1.1 & 0.3 & 0.4 & 1.1 & 0.5 & 7.6 & 3.4 & 0.4 & 0.6 & 3.6 & 0.0 & 3.4 & 0.5 & 4.5 & 4.1 & 0.1 & 1.3 & 1.7
    \\
    
       && & \textsc{S}
    & 7.9 & 35.8 & 15.9 & 17.2 & 13.3 & 4.5 & 2.7 & 1.6 & 5.4 & 2.4 & 15.8 & 7.3 & 4.1 & 2.5 & 8.2 & 0.0 & 6.6 & 2.5 & 8.6 & 7.8 & 0.0 & 6.4 & 5.4
    \\
    
        && & \textsc{M}
    & 1.5 & 7.5 & 3.3 & 2.9 & 2.9 & 0.6 & 0.3 & 0.4 & 0.2 & 0.3 & 3.6 & 1.1 & 0.1 & 0.5 & 2.6 & 0.0 & 1.5 & 0.2 & 1.5 & 2.0 & 0.1 & 1.0 & 0.6
    \\
     \cmidrule{2-27}
    

    & \multirow{6}{*}{\sc 39}
    & \multirow{3}{*}{\rotatebox[origin=c]{90}{\textsc{ind}}} 
    & \textsc{A}
    
    & 6.1 & 15.6 & 9.1 & 11.5 & 10.5 & 2.8 & 6.7 & 3.7 & 3.2 & 6.7 & 13.2 & 7.9 & 4.0 & 4.6 & 6.7 & 0.9 & 4.3 & 2.1 & 7.4 & 7.2 & 0.0 & 2.3 & 3.3 \\ 
    
        && & \textsc{S}
    & 19.4 & 36.4 & 24.1 & 30.3 & 25.6 & 14.3 & 18.5 & 34.7 & 12.2 & 31.5 & 23.4 & 36.0 & 29.8 & 17.8 & 18.5 & 6.1 & 8.5 & 5.0 & 16.9 & 21.3 & 0.0 & 5.4 & 9.3 \\ 
    
        && & \textsc{M}
    & 3.0 & 7.2 & 3.9 & 4.9 & 4.6 & 1.5 & 6.3 & 2.6 & 1.0 & 2.9 & 7.3 & 3.9 & 1.5 & 4.1 & 2.5 & 0.0 & 2.2 & 1.4 & 1.8 & 3.3 & 0.0 & 1.4 & 0.6
    \\ \cmidrule{3-27}
    
    && \multirow{3}{*}{\rotatebox[origin=c]{90}{\textsc{conf}}} 
       &  \textsc{A}
    & 4.9 & 12.1 & 8.2 & 9.6 & 8.8 & 2.4 & 3.1 & 3.3 & 2.9 & 5.9 & 9.3 & 7.4 & 3.5 & 1.9 & 5.6 & 0.8 & 4.1 & 1.7 & 6.8 & 5.7 & 0.0 & 1.8 & 3.3
    \\
    
       && & \textsc{S}
    &17.4 & 32.6 & 22.9 & 26.5 & 23.4 & 12.2 & 16.7 & 32.4 & 11.2 & 28.3 & 19.3 & 34.3 & 27.1 & 15.9 & 16.0 & 5.6 & 8.2 & 4.7 & 14.9 & 17.2 & 0.0 & 5.1 & 9.2
    \\
    
        && & \textsc{M}
    & 2.1 & 4.6 & 3.3 & 3.6 & 3.0 & 1.2 & 2.6 & 2.3 & 0.8 & 2.7 & 3.9 & 3.7 & 1.4 & 1.5 & 1.8 & 0.0 & 2.1 & 1.0 & 1.9 & 2.0 & 0.0 & 1.0 & 0.7
    \\
    \cmidrule{2-27}
    
 
    & \multirow{6}{*}{\sc 93}
    & \multirow{3}{*}{\rotatebox[origin=c]{90}{\textsc{ind}}} 
    & \textsc{A}
    & 5.8 & 13.9 & 7.6 & 10.1 & 11.2 & 2.8 & 7.2 & 2.9 & 2.9 & 5.8 & 13.6 & 8.1 & 4.4 & 3.2 & 7.2 & 0.6 & 3.1 & 2.4 & 6.8 & 6.6 & 1.0 & 2.5 & 3.2 \\ 
    
        && & \textsc{S}
    & 18.5 & 34.5 & 20.0 & 28.9 & 25.3 & 14.0 & 20.1 & 26.0 & 13.0 & 28.6 & 25.4 & 35.0 & 25.6 & 17.3 & 18.2 & 4.9 & 6.7 & 7.2 & 13.6 & 17.6 & 11.6 & 5.7 & 8.3 \\ 
    
        && & \textsc{M}
    & 2.7 & 6.6 & 3.0 & 4.7 & 5.2 & 1.3 & 6.8 & 2.3 & 0.9 & 2.5 & 7.6 & 4.3 & 2.1 & 2.7 & 3.1 & 0.0 & 1.8 & 0.9 & 1.3 & 1.4 & 0.2 & 1.3 & 0.4
    \\ \cmidrule{3-27}
    
    && \multirow{3}{*}{\rotatebox[origin=c]{90}{\textsc{conf}}} 
       &  \textsc{A}
    & 4.6 & 11.3 & 6.4 & 8.6 & 9.1 & 2.2 & 2.7 & 2.5 & 2.7 & 4.9 & 10.5 & 7.2 & 3.5 & 1.8 & 6.1 & 0.6 & 2.8 & 2.0 & 6.2 & 5.1 & 0.8 & 1.4 & 2.5
    \\
    
       && & \textsc{S}
    & 16.3 & 31.5 & 18.4 & 26.3 & 22.3 & 11.8 & 18.0 & 24.3 & 12.2 & 25.5 & 22.3 & 31.3 & 20.8 & 14.5 & 16.4 & 4.4 & 6.1 & 6.6 & 11.6 & 15.3 & 8.9 & 2.4 & 7.6
    \\
    
        && & \textsc{M}
    & 1.8 & 4.7 & 2.1 & 3.5 & 3.4 & 0.9 & 2.3 & 2.0 & 0.8 & 2.1 & 4.9 & 3.5 & 1.5 & 1.4 & 2.2 & 0.0 & 1.6 & 0.7 & 1.1 & 0.9 & 0.1 & 0.6 & 0.4
    \\
    \midrule

    \multirow{6}{*}{\rotatebox[origin=c]{90}{\textsc{pep\textsubscript{ms}}}}  & \multirow{6}{*}{\sc 39}
    & \multirow{3}{*}{\rotatebox[origin=c]{90}{\textsc{ind}}} 
    & \textsc{A}
    & 4.7 & 15.1 & 6.9 & 11.0 & 9.6 & 5.0 & 3.8 & 3.2 & 2.0 & 7.3 & 9.0 & 5.5 & 3.0 & 3.3 & 1.9 & 0.2 & 3.3 & 1.5 & 5.9 & 5.5 & 0.0 & 0.7 & 0.6
    \\
    
        && & \textsc{S}
    & 15.0 & 35.2 & 18.6 & 29.4 & 22.0 & 16.7 & 15.7 & 19.4 & 8.7 & 29.3 & 19.2 & 30.2 & 24.5 & 19.9 & 4.6 & 1.7 & 6.4 & 2.5 & 10.3 & 12.8 & 0.0 & 1.1 & 1.2
    \\
    
        && & \textsc{M}
    & 2.4 & 7.1 & 2.4 & 4.5 & 4.2 & 2.1 & 3.5 & 2.6 & 0.7 & 3.3 & 4.4 & 3.5 & 0.5 & 2.7 & 1.0 & 0.0 & 2.0 & 1.2 & 2.4 & 2.8 & 0.0 & 0.5 & 0.4
    \\ 
    \cmidrule{3-27}
    
    && \multirow{3}{*}{\rotatebox[origin=c]{90}{\textsc{conf}}} 
      &  \textsc{A}
    &6.0 & 15.7 & 8.1 & 12.5 & 11.7 & 5.7 & 6.9 & 5.2 & 2.9 & 9.2 & 14.0 & 6.3 & 5.1 & 6.7 & 3.4 & 0.4 & 3.4 & 1.5 & 6.4 & 5.6 & 0.0 & 0.5 & 0.5
    \\
    
        && & \textsc{S}
    & 13.1 & 31.9 & 17.1 & 27.1 & 19.6 & 12.1 & 13.6 & 17.7 & 7.6 & 26.1 & 16.0 & 27.1 & 21.4 & 16.9 & 3.9 & 1.6 & 5.3 & 2.2 & 9.4 & 9.0 & 0.0 & 0.8 & 1.1
    \\
    
        && & \textsc{M}
    &  4.3 & 10.0 & 4.5 & 7.1 & 8.6 & 4.0 & 6.7 & 4.7 & 2.0 & 6.2 & 11.8 & 4.9 & 3.3 & 6.4 & 2.7 & 0.2 & 2.4 & 1.3 & 3.9 & 3.8 & 0.0 & 0.2 & 0.3
    \\ \midrule
    
    \multirow{18}{*}{\rotatebox[origin=c]{90}{\textsc{pep\textsubscript{ms}+mlm}}}  & \multirow{6}{*}{\sc en}
    & \multirow{3}{*}{\rotatebox[origin=c]{90}{\textsc{ind}}} 
    & \textsc{A}
    & 2.6 & 16.8 & 5.0 & 5.2 & 4.9 & 1.5 & 0.2 & 0.6 & 0.2 & 0.6 & 6.3 & 3.0 & 0.4 & 0.6 & 1.0 & 0.0 & 1.2 & 0.4 & 4.5 & 2.3 & 0.6 & 1.8 & 0.5 \\ 
    
        && & \textsc{S}
    & 6.5 & 35.5 & 13.4 & 15.5 & 12.3 & 6.2 & 2.0 & 4.7 & 0.5 & 3.3 & 13.0 & 8.7 & 3.1 & 3.0 & 1.9 & 0.0 & 2.5 & 0.5 & 7.0 & 5.3 & 0.7 & 4.5 & 0.4\\ 
    
        && & \textsc{M}
    & 1.2 & 6.6 & 2.2 & 2.7 & 2.3 & 1.2 & 0.2 & 0.5 & 0.1 & 0.3 & 2.8 & 0.7 & 0.1 & 0.5 & 0.3 & 0.0 & 0.6 & 0.4 & 1.4 & 1.0 & 0.5 & 1.3 & 0.5
    \\ \cmidrule{3-27}
    
    && \multirow{3}{*}{\rotatebox[origin=c]{90}{\textsc{conf}}} 
       &  \textsc{A}
    & 2.7 & 18.0 & 5.3 & 5.1 & 4.6 & 1.5 & 0.6 & 1.0 & 0.2 & 0.9 & 6.3 & 3.0 & 0.6 & 0.9 & 0.9 & 0.0 & 1.5 & 0.3 & 4.6 & 2.1 & 0.6 & 1.6 & 0.4
    \\
    
       && & \textsc{S}
    & 5.7 & 33.1 & 12.0 & 13.3 & 10.2 & 5.8 & 2.0 & 4.4 & 0.5 & 2.7 & 11.1 & 7.3 & 3.1 & 3.0 & 1.7 & 0.0 & 1.7 & 0.3 & 6.7 & 4.6 & 0.6 & 1.0 & 0.3
    \\
    
        && & \textsc{M}
    & 1.6 & 10.4 & 3.0 & 3.2 & 2.6 & 1.2 & 0.5 & 0.9 & 0.1 & 0.8 & 3.8 & 1.2 & 0.3 & 0.8 & 0.3 & 0.0 & 1.1 & 0.3 & 1.8 & 1.1 & 0.5 & 1.2 & 0.5
    \\
     \cmidrule{2-27}
    

    & \multirow{6}{*}{\sc 39}
    & \multirow{3}{*}{\rotatebox[origin=c]{90}{\textsc{ind}}} 
    & \textsc{A}
    & 4.9 & 14.9 & 9.7 & 10.5 & 10.5 & 7.3 & 5.5 & 4.4 & 1.3 & 7.0 & 9.4 & 5.8 & 2.0 & 2.6 & 1.6 & 0.0 & 2.8 & 0.9 & 4.5 & 6.5 & 0.0 & 0.3 & 0.4 \\ 
    
        && & \textsc{S}
    & 13.9 & 34.8 & 24.2 & 27.7 & 23.1 & 20.2 & 16.0 & 17.5 & 5.8 & 28.6 & 19.3 & 25.2 & 13.5 & 15.3 & 4.0 & 2.4 & 5.5 & 1.7 & 8.3 & 11.2 & 0.0 & 0.3 & 0.7\\ 
    
        && & \textsc{M}
    &2.4 & 6.3 & 3.4 & 4.6 & 4.2 & 3.8 & 5.2 & 2.7 & 0.5 & 3.0 & 4.6 & 4.1 & 0.6 & 2.1 & 0.7 & 0.0 & 1.7 & 0.8 & 1.9 & 3.1 & 0.0 & 0.3 & 0.2
    \\ \cmidrule{3-27}
    
    && \multirow{3}{*}{\rotatebox[origin=c]{90}{\textsc{conf}}} 
       &  \textsc{A}
    & 5.7 & 16.3 & 10.5 & 11.0 & 11.3 & 7.9 & 7.0 & 6.2 & 1.3 & 9.2 & 14.2 & 6.6 & 2.8 & 4.2 & 1.5 & 0.3 & 3.4 & 1.3 & 4.1 & 6.6 & 0.0 & 0.3 & 0.3
    \\
    
       && & \textsc{S}
    & 12.0 & 30.6 & 21.6 & 24.8 & 20.1 & 16.6 & 15.7 & 16.0 & 3.0 & 27.1 & 15.8 & 20.8 & 12.3 & 12.8 & 3.4 & 0.0 & 4.9 & 2.0 & 7.0 & 9.5 & 0.0 & 0.2 & 0.4
    \\
    
        && & \textsc{M}
    & 3.9 & 9.4 & 5.4 & 6.2 & 6.7 & 5.0 & 6.7 & 4.5 & 0.8 & 6.0 & 11.8 & 5.6 & 1.5 & 3.8 & 0.7 & 0.3 & 2.8 & 1.3 & 2.6 & 4.5 & 0.0 & 0.3 & 0.2
    \\
    \cmidrule{2-27}
    
 
    & \multirow{6}{*}{\sc 93}
    & \multirow{3}{*}{\rotatebox[origin=c]{90}{\textsc{ind}}} 
    & \textsc{A}
    & 4.5 & 14.1 & 8.6 & 9.1 & 8.8 & 5.8 & 5.0 & 3.1 & 0.8 & 6.3 & 9.4 & 5.7 & 1.8 & 1.5 & 1.5 & 0.1 & 2.9 & 1.7 & 4.0 & 5.2 & 0.7 & 2.3 & 0.4 \\ 
    
        && & \textsc{S}
    &13.2 & 32.7 & 22.0 & 24.4 & 20.3 & 18.6 & 17.0 & 12.8 & 3.8 & 26.9 & 19.4 & 25.9 & 12.2 & 12.0 & 3.4 & 0.9 & 5.0 & 2.1 & 9.7 & 9.5 & 8.1 & 3.1 & 0.9\\ 
    
        && & \textsc{M}
    & 2.3 & 5.9 & 3.3 & 4.4 & 3.8 & 3.1 & 4.7 & 2.6 & 0.4 & 2.8 & 4.4 & 3.8 & 0.4 & 1.1 & 0.7 & 0.0 & 2.2 & 1.5 & 1.2 & 1.3 & 0.0 & 1.9 & 0.2
    \\ \cmidrule{3-27}
    
    && \multirow{3}{*}{\rotatebox[origin=c]{90}{\textsc{conf}}} 
       &  \textsc{A}
    &5.5 & 15.7 & 9.5 & 9.8 & 9.3 & 6.2 & 6.9 & 5.0 & 0.8 & 8.1 & 13.5 & 7.3 & 2.8 & 3.0 & 1.4 & 0.1 & 2.6 & 1.8 & 4.4 & 6.8 & 2.0 & 2.6 & 0.4
    \\
    
       && & \textsc{S}
    & 11.9 & 31.3 & 19.2 & 22.0 & 16.8 & 15.7 & 16.3 & 12.5 & 3.5 & 25.0 & 16.6 & 23.9 & 11.3 & 8.6 & 2.8 & 0.6 & 4.3 & 2.0 & 9.0 & 9.4 & 7.5 & 2.7 & 0.8
    \\
    
        && & \textsc{M}
    & 3.8 & 8.7 & 5.2 & 6.0 & 5.5 & 4.2 & 6.6 & 4.5 & 0.5 & 5.4 & 10.9 & 5.9 & 1.7 & 2.7 & 0.7 & 0.1 & 2.3 & 1.8 & 2.3 & 3.7 & 1.5 & 2.6 & 0.2
    \\
    
    \bottomrule
    \end{tabular}
    }
    \caption{X-FACTR results.}
    \label{tab:xfactr_results}
\end{table*}

\end{document}